\documentclass{article}
\usepackage{jfrExamplee}
\usepackage{graphicx}
\usepackage{apalike}
\usepackage{setspace}

\usepackage{enumitem} 

\usepackage{hyperref}
\usepackage{amsmath,amsfonts}
\usepackage{algorithm,algpseudocode}
\usepackage{booktabs}
\usepackage{color}
\usepackage{cite}
\usepackage{amssymb}
\usepackage{subfigure}
\usepackage{array}

\usepackage{acronym}
\acrodef{GESTALT}{Grid-based Estimation of Surface Traversability Applied to Local Terrain}
\acrodef{JPL}{Jet Propulsion Laboratory}
\acrodef{VO}{Visual Odometry}
\acrodef{MSL}{Mars Science Laboratory}
\acrodef{MER}{Mars Exploration Rover}
\acrodef{POMDP}{Partially Observable Markov Decision Process}
\acrodef{ROAMS}{Rover Analysis Modeling and Simulation}
\acrodef{MAV}{Micro Aerial Vehicle}
\acrodef{SLAM}{Simultaneous Localization and Mapping}
\acrodef{RRBT}{Rapidly-exploring Random Belief Tree}
\acrodef{RHC}{Receding Horizon Control}
\acrodef{FOV}{Field of View}
\acrodef{DLT}{Direct Linear Transform}
\acrodef{ROS}{Robot Operating System}
\acrodef{FSM}{finite-state machine}
\acrodef{sol}{Martian solar day}
\acrodef{MSR}{Mars Sample Return}
\acrodef{SOG}{Sum of Gaussians}
\acrodef{UAV}{Unmanned Aerial Vehicle}
\acrodef{AUV}{Autonomous Underwater Vehicle}
\acrodef{RRT}{Rapidly-exploring Random Trees}
\acrodef{PRM}{probabilistic roadmap}
\acrodef{RRBT}{Rapidly-exploring Random Belief Trees}
\acrodef{BRM}{belief roadmap}
\acrodef{GPS}{Global Positioning System}
\acrodef{GNSS}{Global Navigation Satellite System}
\acrodef{DSN}{Deep Space Network}

\newcolumntype{M}[1]{>{\centering\arraybackslash}m{#1}}
\newcommand{\myvala}{0.2}

\newcommand{\myvalc}{0.23}
\newcommand{\myvald}{3.7cm}
\newcommand{\rev}[1]{\textcolor{black}{#1}} 

\title{Perception-aware Autonomous Mast Motion Planning for Planetary Exploration Rovers : preprint version}

\author{
Jared ~Strader\\
Department of Mechanical and Aerospace Engineering \\
West Virginia University\\
Morgantown, WV 26506 \\
\And
Kyohei Otsu and Ali-akbar Agha-mohammadi \\
Jet Propulsion Laboratory\\
California Institute of Technology \\
Pasadena, CA 91109 \\
}

\begin{document}

\maketitle

\begin{abstract}

Highly accurate real-time localization is of fundamental importance for the safety and efficiency of planetary rovers exploring the surface of Mars. 
Mars rover operations rely on vision-based systems to avoid hazards as well as plan safe routes. 
However, vision-based systems operate on the assumption that sufficient visual texture is visible in the scene. 
This poses a challenge for vision-based navigation on Mars where regions lacking visual texture are prevalent. 
To overcome this, we make use of the ability of the rover to actively steer the visual sensor to improve fault tolerance and maximize the perception performance.
This paper answers the question of \textit{where and when} to look by presenting a method for predicting the sensor trajectory that maximizes the localization performance of the rover. 
This is accomplished by an online assessment of possible trajectories using synthetic, future camera views created from previous observations of the scene.
The proposed trajectories are quantified and chosen based on the expected localization performance. 
\rev{In this work, we validate the proposed method in field experiments at the Jet Propulsion Laboratory (JPL) Mars Yard. Furthermore, multiple performance metrics are identified and evaluated for reducing the overall runtime of the algorithm.}
We show how actively steering the perception system increases the localization accuracy compared to traditional fixed-sensor configurations.
\end{abstract}


\section{Introduction}

\subsection{Motivating Example}

Highly accurate real-time localization is of fundamental importance for both the safety and efficiency of planetary rovers exploring the surface of Mars. 
The position of the rover is crucial for identifying and avoiding hazards as well as manually and autonomously planning safe routes. 
Recent generations of Mars rovers each use a similar approach for localization. While driving, the attitude angles are propagated using gyroscopes, and the position is propagated using wheel odometry \cite{grotzinger2012mars}. 
However, the pose suffers from drift due to wheel slip, and as a result, the rover often displays large localization error in the presence of large rocks, steep slopes, and sandy terrain. 
To mitigate errors, \ac{VO} is used to refine the pose of the rover, and now, vision-based localization is a key technology used for Mars rover operations \cite{Maimone07,johnson2008}. 
This poses several challenges for future Mars missions where more self-reliant rovers are required for accurate, safe, and efficient autonomous navigation.

\begin{figure*}[h]
\centerline
{
	\subfigure[Perseverance Valley located in Meridiani Planum recorded by NASA's Opportunity Mars rover. In parts of this region, sufficient visual texture for navigation is difficult to find nearby the rover.]
	{
		\includegraphics[width=0.47\columnwidth]{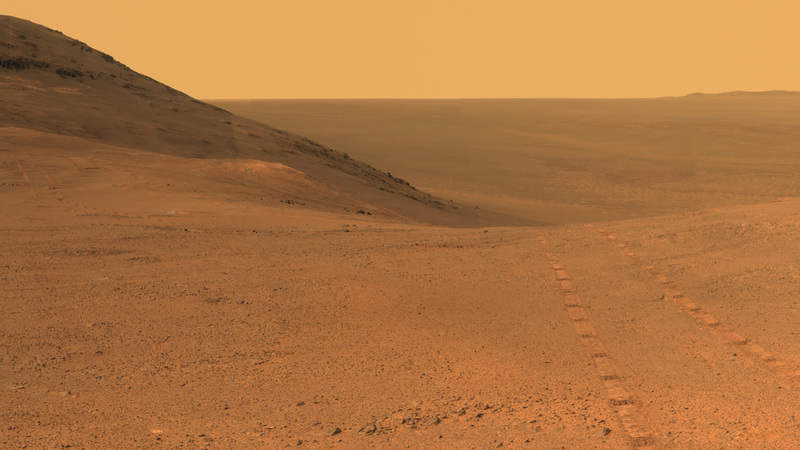}
		\label{fig:meridiani}
	}
	\hspace{0.5\baselineskip}
	\subfigure[Self-portrait of NASA's Curiosity Mars rover after reaching Namib Dune. Different mast camera directions lead to significantly different VO accuracy as part of the terrain is feature-poor (sandy) and the other part is feature-rich (rocks).]
	{
		\includegraphics[width=0.47\columnwidth]{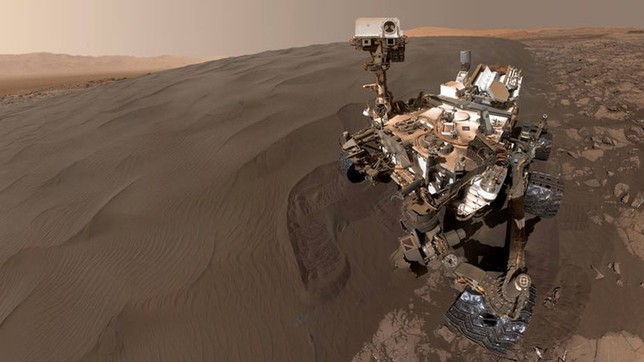}
		\label{fig:rover}
	}
}
\caption{Examples of challenging environments for vision-based navigation on Mars.}
\label{fig:illustration}
\end{figure*}

In general, vision-based systems operate on the assumption that the terrain contains sufficient visual texture for navigation. 
This poses a challenge on Mars where regions lacking visual texture are prevalent. 
For example, consider the terrain at Meridiani Planum (pictured in Figure \ref{fig:meridiani}) explored by Opportunity during the \ac{MER} mission; in parts of this region, sufficient visual texture for navigation is difficult or nearly impossible to find nearby the rover. 
This is caused by a thick layer of regolith covering the ground. Fortunately, Meridiani Planum is mostly flat, so the rover only suffers from minimal localization drift; however, in the future, regions of scientific interest may not provide terrain as mild as seen here. 
On the 1228th \ac{sol} of the \ac{MSL} mission, the Curiosity rover reached Namib Dune (pictured in Figure \ref{fig:rover}). Namib Dune is part of the vast dark-sand area referred to as the "Bagnold Dune Field" located along the northwestern flank of Mount Sharp on Mars. 
The large sand dunes found in this region are difficult for vision-based systems due to the lack of visual texture.
In these regions, the sparsity of visual texture can degrade the perception performance leading to increased localization drift.
In some cases, this results in total \ac{VO} failure. For example, on the 235th \ac{sol} of the \ac{MER} mission, one of the few instances of \ac{VO} failure occurred due to the lack of useful visual features extracted from the viewpoint of the camera \cite{Maimone07}.

To avoid \ac{VO} performance degradation on current Mars missions, rover operators on Earth manually point the mast in order to maximize the visual texture (e.g. quality and quantity of features) visible from the viewpoint of the camera. 
This has been the regular mode of operation in scenes lacking visual textures. 
However, manual pointing of the mast is tedious and requires frequent stops. This limits rover autonomy, and consequently, the total distance the rover is capable of traversing is reduced per \ac{sol}. 
This mode of operation is not adequate for future Mars missions as they will require increased operational efficiency in comparison to previous missions. 
For example, the ``fetch'' rover of the \ac{MSR} mission is primarily drive-based and will be required to drive more than ten kilometers within a few months \cite{Mattingly2011}.
The current Mars rovers drive slowly due to factors such as science objectives, communication constraints, and limited autonomy. 
Reliable navigation will significantly improve autonomous capabilities and reduce the frequency of manual commands required to operate the rover. 

\rev{In general, texture limited surfaces are not specific to space robotics but exist on terrestrial environments as well. 
For example, arctic and volcanic regions are some of the harshest environments on Earth and suffer many challenges similar to that of systems on Mars. 
In recent years, mobile robots have been deployed in such harsh environments for scientific studies \cite{ray2007design,muscato2012volcanic} as well as for aiding in natural disasters \cite{murphy2016disaster}. Several works have addressed some of the issues regarding vision-based systems in these harsh environments \cite{williams2010developing,otsu2015experiments}. 
However, space robotics are particularly difficult in regards to navigation since Earth-based technologies such as \acp{GNSS} and magnetic compassing are typically not applicable in space. 
Additionally, systems developed for space applications are required to operate under strict communication requirements due to limitations such as availability of the \ac{DSN}, onboard technologies, and environmental constraints (e.g., signal delay and signal blockage). 
The computational resources in space robotics are severely limited in comparison to modern computers due to radiation-hardened hardware, and as a result, the onboard capabilities are typically limited in comparison to most Earth-based systems.}


\subsection{Contribution}

\rev{To address the described challenges, we employ} a method that automates the process of pointing the mast in order to maximize the useful visual information in the viewpoint of the camera. 
We make use of the ability of the rover to actively steer the visual sensor to improve fault tolerance (e.g. \ac{VO} failure) and maximize perception performance (e.g., \rev{minimize} VO drift).
This problem is an instance of the general perception-aware planning problem in robotics with applications including ground, air, and underwater vehicles as well as grasping and manipulation to name a few. In general, we address the problems of 1) ``What to measure?" and 2) ``How to measure?" in order to improve the performance of the estimator under the constraints imposed by the sensor capabilities and computational resources. 
In the most general form, this problem can be cast as a \ac{POMDP}, or belief-space planning problem. 

In our previous work \cite{Otsu18_wheretolook}, we develop basic theory supporting our method and validate the algorithm using a physics-based simulator \ac{ROAMS} to show improvement in performance compared to the traditional fixed-mast scenario. 
\rev{Specifically, we proposed a foward solution for belief-space planning using motion primitives designed for planning mast motions, and we assessed candidate motions over the search space using predictive perception models to quantify visual odometry performance.} 
However, while the simulator was adequate for validating the general theory, the effectiveness and practicality of the algorithm has yet to be verified in a realistic environment. Therefore, in this work, we generalize and expand our previous work and demonstrate the effectiveness of our method on actual hardware in a Mars-analogue environment.


\rev{Our contributions are outlined as follows: 
1)  We provide a general framework for perception-aware planning agnostic to the specific algorithm used for \ac{VO}. 
This is achieved by directly minimizing \ac{VO} error in contrast to minimizing algorithm dependent metrics. 
2) We decouple path planning and observation planning by actuating the visual sensors, which allow for efficient planning during runtime. 
To achieve this, a spatio-temporal variant of \ac{RRT} is employed, and to the best of our knowledge, our method is the first consider this type of decoupling for belief space planning. 
3) We validate our method on actual hardware in a Mars-analogue environment. 
In this respect, we provide a detailed description of our implementation and hardware specifications, so the reader can more easily replicate our work. 
4) We compare and contrast a set of additional performance metrics (in addition to \ac{VO} error) to reduce the overall runtime of the algorithm. 
The additional performance metrics proposed in this work are valid for sparse methods for \ac{VO} (in contrast to direct methods).}



\subsection{Outline of Paper}
The remainder of this paper is structured as follows: In Section \ref{sec:RW}, we review the literature on related works. In Section \ref{sec:problem}, the problem is formally defined with the relevant mathematical notation. The method for mapping and information representation is described in Section \ref{sec:method}. The problem and approach for predictive perception is described in Section \ref{sec:predictivePerception}, and Section \ref{sec:planning} covers the planning methods. We provide additional implementation details for field experiments in Section \ref{sec:implementation}, and we discuss the field experiments and the accuracy and efficiency of our method in Section \ref{sec:field_experiments}. The paper is concluded in Section \ref{sec:conclusion}.


\section{Related Work}

\label{sec:RW}
Our work lies in the intersection of perception and planning for planetary rovers, active vision, and planning under uncertainty. Thus, we review the related literature in these domains.

\subsection{Perception and Planning for Planetary Rovers}
While Mars rover operations require frequent commands from human operators, rovers still rely on autonomous navigation frameworks to ensure the safety of the rover while exploring the surface of Mars. 
These frameworks increase the efficiency of rover operations by allowing operators to provide high-level commands to the rover. The current frameworks are primarily vision-based and rely on passive cameras for perception.

Currently, Mars rover missions utilize the \ac{GESTALT} navigation system  \cite{goldberg-2002,biesiadecki2006mars} for path planning to enable the rover to safely traverse short distances using stereo vision. 
This system is based on the earlier Morphin algorithm \cite{simmons2001} and allows the rover to autonomously detect hazardous terrain nearby the rover. 
This is achieved by building a traversablity map using the range data generated from the onboard stereo system. 
The traversability map is built by discretizing the world into a grid such that each grid cell is roughly the size of one of the rover's wheels. 
Each of the cells are assigned values describing traversability such as slope, roughness, elevation, and the traversability of neighboring cells.
The resulting map is used to plan the safest route to the desired goal.
The rover does not look for obstacles while driving using the \ac{GESTALT} system, but other safeguards are typically enabled to ensure the safety of the rover (e.g. tilt sensors and motor current limits).

As the rover traverses the resulting trajectories, the rover utilizes stereo \acf{VO} to refine the pose of the rover with respect to the generated map. 
One of the earliest works on the estimation of robot motion with stereo cameras (i.e. stereo \ac{VO}) goes back to \cite{Moravec:1980:OAN:909315}. 
Along this line of work, \cite{matthies1987error} proposed the first quantitatively accurate stereo vision-based egomotion estimation results, which led to the \ac{VO} algorithm currently in use on Mars. Some minor modifications have been developed to improve the robustness and accuracy of the algorithm \cite{olson2003rover}. 
The general approach to estimate motion using stereo \ac{VO} is to find features in a stereo image pair and track them between consecutive frames. 
The general elements of the algorithm originally formulated in \cite{Moravec:1980:OAN:909315} are the following: i) image de-warping and rectification, ii) feature detection, iii) feature triangulation, iv) feature tracking, v) outlier rejection, and vi) nonlinear refinement. 
Sparse \ac{VO} was adapted specifically for stereo cameras by Matthies \cite{Matthies:1989:DSV:916891}, and later integrated and successfully deployed on the \ac{JPL} rover platforms, \ac{MER}\cite{Maimone07} and \ac{MSL}\cite{johnson2008}.

In general, vision-based navigation systems operate on the assumption that the environment contains sufficient visual texture for navigation. 
\rev{This poses a problem for most vision-based navigation systems. Vision-based perception for planetary rovers must be capable of dealing with challenging environments containing texture-poor terrain (e.g. sandy dunes).}
In some cases, the wheel tracks provide enough features to estimate the motion of the rover using stereo \ac{VO}, which has been verified in \cite{Maimone07}.
However, relying on wheel tracks for autonomous operations risk potential inaccuracies, and as a result, rover operators provide frequent low-level commands such as pointing of the mast to avoid these challenging situations.
\rev{This requires operating the rover only during short periods where communication with the rover is possible.}
In this case, active perception can be used to provide a more self-reliant navigation system, which is able to actively avoid potential inaccuracies in navigation.

\subsection{Planning Under Uncertainty} 
Perception-aware autonomous mast motion planning is an instance of sequential decision making under uncertainty. 
In its most general form, Partially Observable Markov Decision Processes (POMDPs) \cite{Kaelbling98} provide a general and principled framework for the problem of sequential decision making under uncertainty. While only a narrow class of problems formulated as POMDPs can be solved exactly due to the computational complexity \cite{Madani99,Papadimitriou87}, a strong body of literature exists on techniques for approximating POMDPs. The literature on general purpose POMDPs can be separated into two main categories: offline solvers and online search algorithms. 

The first category, offline solvers, computes a policy over the belief space, or in other words, offline solvers aim to find an action for all possible beliefs \cite{Pineau03,Spaan05,Smith05-HSVI,Ali14-IJRR}. 
The second category, online search algorithms, aims to find the best action for the current belief using a forward search tree rooted in the current belief \cite{Ross_2007_AEMS,ye2017despot,kurniawati2016online,silver2010monte}. 
In the case real-time replanning is required due to limited prior information, a \ac{RHC} is often used to reduce computational complexity. 
For replaninng purposes, forward search algorithms can be used in a \ac{RHC} scheme. 
Recently, methods using \ac{RHC} have been extended for the belief space as well as dynamic environments \cite{agha2018slap,kim2019bi,Erez2010,Platt10,Chakrav11-IRHC,He11JAIR,Toit10}. 
In an \ac{RHC} scheme, optimization is performed only within a limited horizon; thus, the system performs optimization within the specified horizon, then the system takes the next immediate action and moves the optimization horizon one step forward before repeating the process. 

Probabilistic sampling-based algorithms are often used with great success in motion planning problems in robotics. In general, probabilistic sampling-based algorithms can be divided into multiple-query methods such as \ac{PRM} \cite{Kavraki96,Kavraki98} and variants \cite{Amato98,bohlin00lazyPRM,Karaman.Frazzoli:IJRR11} and single-query methods \rev{such as \ac{RRT} \cite{lavalle1998rapidly} and variants \cite{kuffner2000rrt,Karaman.Frazzoli:IJRR11,alterovitz2011rapidly,karaman2009sampling}. Extensions of the previous mentioned single-query and multiple-query methods were extended to include both motion uncertainty \cite{alterovitz2007stochastic} as well as sensing uncertainty \cite{Prentice09,Ali-FIRM-ICRA14,Bry11}}.
While multiple-query methods generally perform well in high-dimensional state spaces, single-query methods are useful for online planning due to the incremental nature of the algorithm by avoiding the necessity to initially construct a graph of the configuration space (or roadmap). 
In \rev{\cite{plaku2005sampling} and \cite{janson2015fast}}, the main elements are combined of both single-query and multiple-query algorithms. 
The key to all sampling-based methods is the avoidance of the explicit construction of the configuration space by probing the configuration space with a sampling scheme. 
These methods provide several useful properties such as probabilistic completeness as well as the exponential rate of decay of the probability of failure with the number of samples. 
A detailed analysis of the convergence and optimality for multiple-query and single-query algorithms are provided in \cite{Kavraki98,Karaman.Frazzoli:IJRR11}.

\subsection{Active Perception}
\rev{While the previous mentioned methods incorporate uncertainty in the planning algorithm, few explicitly consider perceptual information as a planning metric. In recent years, several works have emerged that focus on this aspect, which aim to improve perception performance by finding trajectories that not only avoid difficult situations but are optimal in regards to perception. This is often referred to as active perception.}
The main goal of active perception is to optimize sensor placement in order to maximize the performance of a particular task. 
This problem was originally addressed in \cite{aloimonos1988active,bajcsy1988active} and is similar to the sensor planning problem \cite{tarabanis1995survey}. 

One of the goals of active perception is the minimization of pose uncertainty by planning the optimal sensor trajectory. This problem is referred to as active localization and is typically addressed in the context of exploration problems as well as \ac{SLAM}. In some cases, active localization is applied to ensure the observability of system states due to limited sensing (e.g. range-only \cite{Vallicrosa2014,strader2016cooperative} or bearing-only \cite{vander2014cautious} scenarios). For example, in \cite{Vallicrosa2014}, visual contact is achieved for autonomous docking of an \ac{AUV} by ensuring the observability and convergence of the position of a beacon using a \ac{SOG} filter. Similarly, in \cite{strader2016cooperative}, cooperative maneuvers are performed to enable observability and uniquely solve for the relative pose of \rev{a pair} of \acp{UAV}. 
Additionally, active mapping \cite{agha2017grid3d,agha2019confidence,Heiden17,forster2014appearance}, which is often addressed along with active localization, aims to enhance the quality of the map by picking trajectories that maximize the information captured by the sensors. The topic of active perception has been investigated extensively for both range-based sensors \cite{feder1999adaptive,bourgault2002information,vidal2010action,bachrach2012-IJRR} and vision-based sensors \cite{davison2002simultaneous,mostegel2014active,rodrigues2018low,achtelik2014motion,sadat2014feature,Costante2017,vidal2010action,Inoue2016}.
While range-based sensors only observe the geometry of the scene, vision-based sensors observe both texture and geometry providing a richer source of information. 
Here, we limit the remainder of the discussion to vision-based sensors (i.e., active vision) as this is the main focus of our work. 

\rev{One of the earliest works to consider vision-based sensors for active localization is \cite{davison2002simultaneous}. 
This method aims to optimize position uncertainty as well as the integrity and consistency of the map by building a map of robust features for localization purposes.
The goal of this approach is to reduce drift along a predefined trajectory by fixating the visual sensors on particular points.
This method is similar to ours in that the visual sensors are pointed during operation, which allows for observation planning without modifying the path of the robot. However, \cite{davison2002simultaneous} only consider immediate actions instead of sequences of actions like ours. 
Additionally, \cite{mostegel2014active} uses geometric point quality and recognition probability to improve the stability and generation of points in the map.
In contrast to the previous mentioned works, \cite{achtelik2014motion} considers sequences of actions instead of immediate actions. To do this, both \ac{MAV} dynamics and pose uncertainty are incorporated into \ac{RRBT} and demonstrated on a \ac{MAV} performing visual-inertial navigation. While both \cite{davison2002simultaneous} and \cite{mostegel2014active} focus on building high quality maps, \cite{achtelik2014motion} assumes a map is known prior and focuses on planning the best trajectory in terms of localization performance. In contrast, \cite{rodrigues2018low} avoids relying on a map by proposing a lightweight method based only on the current image using artificial potential fields.}

\rev{The works of \cite{sadat2014feature} and \cite{Costante2017} are similar to ours in that each constructs a map during runtime (instead of offline) to plan information-rich trajectories. 
In \cite{mu2015two, mu2016information, mu2017two}, the map is summarized into high-priority features by actively selecting the most informative measurements.
In \cite{sadat2014feature}, a variation of \ac{RRT} is developed to plan a trajectory that maximizes the number of observed features. Similarly, \cite{Costante2017} also developed a variation of \ac{RRT} but instead consider the photometric information (i.e., texture) and geometry to estimate localization uncertainty. While \cite{sadat2014feature} use sparse methods for \ac{VO} and mapping, \cite{Costante2017} use direct methods allowing for improved performance. In contrast, our method is agnostic to the specific algorithm used for \ac{VO} or mapping. This is achieved by directly minimizing \ac{VO} error. Additionally, in our method, path planning is decoupled from observation planning by developing a spatial-temporal variant of \ac{RRT}, which leverages actuated visual sensors. This allows for efficient onboard planning during runtime.}

\section{Problem Description}

\label{sec:problem}
In this section, we provide a formal definition of the addressed problem and summarize the relevant mathematical notation. 

\paragraph{Robot and Mast Description}
Consider the case where a rover is equipped with a movable mast on which cameras are rigidly mounted for navigation. Let the 6 degree of freedom (DoF) pose of the rover be denoted by $\mathbf{x}^{body}$ and the pose of the mast relative to the body be denoted by $\mathbf{x}^{mast}$. The extended state of the rover at the $k$-th time step includes both the pose of the body and mast denoted by $\mathbf{x}_k = \{ \textbf{x}_k^{body}, \textbf{x}_k^{mast} \}$ such that a sequence of poses is defined as $\textbf{x}_{i:j} = \{\textbf{x}_i, \textbf{x}_{i+1}, \cdots, \textbf{x}_j \}$ for $j>i$.

\paragraph{Measurements}
Let an image measurement at the $k$-th time step be denoted by $\textbf{z}_k$ where a sequence of observations is defined as $\textbf{z}_{i:j} = \{ \textbf{z}_i, \textbf{z}_{i+1}, \cdots, \textbf{z}_j \}$ for $j>i$. 

\paragraph{Map Belief}
Let the map of the environment be denoted by $\mathcal{M}$ such that the environment is assumed stationary. The map of the environment is estimated from noisy image measurements; therefore, a probabilistic representation of the map is maintained given by
\begin{align}
    \mathcal{B}_k = p(\mathcal{M}|\mathbf{z}_{0:k},\mathbf{x}_{0:k})
\end{align}
where $\mathcal{B}_k$ is the map belief at the $k$-th time step. In this work, the map is constructed using images captured from the onboard cameras; however, additional image sources could be used such as Mars orbiter imagery or images captured from previous Mars missions. The map representation and update procedure will be described further in Section \ref{sec:mapping}.

\paragraph{Measurement Model}
The measurement model describes the probability distribution over all possible observations given the extended state of the rover $\mathbf{x}$ and the map of the environment $\mathcal{M}$ given by
\begin{align}
    p(\mathbf{z}|\mathcal{M},\mathbf{x}).
\end{align}

\paragraph{Problem Statement}
Given a current map belief, compute the robot extended state trajectory (mast and body states) to the provided goal that maximizes localization accuracy.


\section{Perception-Aware Mast Motion Planning} \label{sec:method}
\begin{figure}[!t]
	\centering
	\includegraphics[width=1\columnwidth]{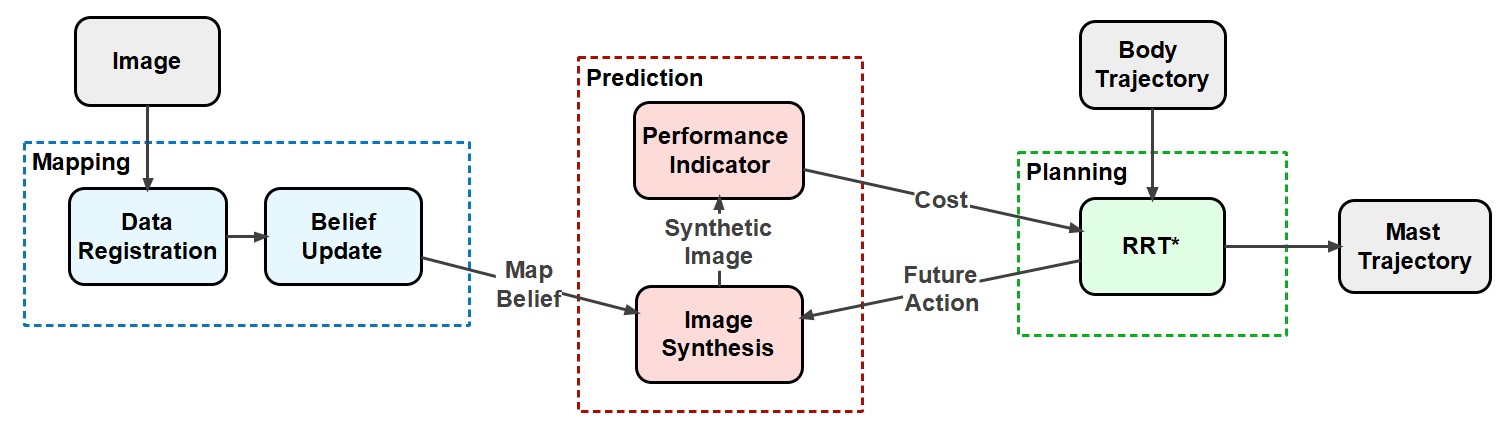}
	\caption{Overview of the proposed method for perception-aware mast planning.}
	\label{fig:algorithm_architecture}
\end{figure}

The proposed method for perception-aware mast planning consists of three primary components: \textit{mapping}, \textit{prediction}, and \textit{planning} as visualized in \autoref{fig:algorithm_architecture}: 
\begin{itemize}
    \item The \textit{mapping} component constructs the map belief $\mathcal{B}_k$ given image measurements $\mathbf{z}_k$ at each time step $k$.
    \item The \textit{prediction} component synthesizes image measurements $\grave{\mathbf{z}}_t$ from the map belief $\mathcal{B}_k$ for potential extended state trajectories $\mathbf{x}_{k:t}$ and computes the perception performance for the predicted measurements.
    \item The \textit{planning} component computes the extended state trajectory of the rover that minimizes the planning objective (e.g. maximizes perception performance). 
\end{itemize}
In this section, we review the theory presented in our previous work \cite{Otsu18_wheretolook}, and we discuss additional implementation details in Section \ref{sec:implementation} to allow the reader to more easily replicate our method.

\subsection{Mapping} \label{sec:mapping}
In this section, we describe the representation of the map and update procedure.

\subsubsection{Map Representation}
In this work, the map is represented as a grid map (3D or 2D) where each voxel contains the visual information of that point in space. Let the map at the $k$-th time step be denoted by $\mathcal{M}_k = \{ \mathbf{m}^{(1)}_k, \textbf{m}^{(2)}_k, \cdots, \textbf{m}^{(N)}_k \}$ where $\mathbf{m}^{(i)}_k$ is the $i$-th voxel in the map. Note the map reduces to a point cloud as the voxel size approaches zero.

Each voxel is characterized by an information pair $\textbf{m}^{(i)}_k = \{I^{(i)}_k, {}^m\mathbf{p}^{(i)}_k \}$ where $I^{(i)}_k$ is the visual information (e.g. intensity) of the $i$-th voxel and ${}^m\mathbf{p}_k^{(i)} = [x^{(i)},y^{(i)},z^{(i)}]^T$ is the geometric coordinates of the center of this voxel in space. Similarly, we denote the geometric coordinates of the $i$-th voxel in the camera references frame as ${}^c\mathbf{p}_k^{(i)}$.

The map at the $k$-th time step is created based on past images, $\mathbf{z}_{0:k}$, and the associated rover states at which the images were acquired, $\mathbf{x}_{0:k}$. Due to measurement noise, the best one can do is create a map belief, $\mathcal{B}_k = p(\mathcal{M}|\mathbf{x}_{0:k},\mathbf{z}_{0:k})$. Instead of maintaining the full map belief, the map is approximated with a collection of beliefs,
\begin{align} \label{eq:independent_assumption}
    b^{(i)}_k = p(\mathbf{m}^{(i)}|\mathbf{x}_{0:k},\mathbf{z}_{0:k}),
\end{align}
where $b^{(i)}_k$ is the belief of the $i$-th voxel such that $\mathcal{B}_k = \{b_k^{(1)}, b_k^{(2)}, \cdots, b_k^{(N)}\}$.

\subsubsection{Map Update}
At each step the rover captures an image $\mathbf{z}_k$, the map is recursively updated by computing the belief at each voxel,
\begin{align} \label{eq:belief_propagation}
    b_k^{(i)} = \tau (b_{k-1}^{(i)}, \mathbf{x}_k, \mathbf{z}_k),
\end{align}
where $\tau$ is the process of belief propagation described in the remainder of this section.



The $i$-th voxel in the camera reference frame ${}^c\mathbf{p}^{(i)}$ can be mapped to the corresponding pixel in the image plane using a perspective projection,
\begin{align} \label{eq:camera_projection_model}
    \mathbf{u}^{(i)} = \mathcal{P}({}^c\mathbf{p}^{(i)}),
\end{align}
where $\mathcal{P}: \mathbb{R}^3 \rightarrow \mathbb{R}^2$ is the camera projection model determined from the intrinisic camera parameters and $\mathbf{u}^{(i)} = [u^{(i)},v^{(i)}]^T$ are the pixel coordinates of the $i$-th voxel. 

To associate a pixel in the image plane to a voxel in the scene, the inverse projection $\mathcal{P}^{-1}$ can be used given the depth,
\begin{align} \label{eq:inv_camera_projection_model}
    {}^c\mathbf{p}^{(i)} = \mathcal{P}^{-1}(\mathbf{u}^{(i)},d_{\mathbf{u}}^{(i)}),
\end{align}
where $d_{\mathbf{u}}^{(i)}$ is the distance to the $i$-th voxel from the viewpoint of the camera. \rev{To update the map, the voxels are transformed from the camera reference frame to the map by
\begin{align} \label{eq:data_registration}
    {}^m\mathbf{p}^{(i)} = {}^m\mathbf{T}_c [ {}^c\mathbf{p}^{(i)}, \; 1]^T
\end{align}
where ${}^m \mathbf{T}_c \in \text{SE}(3)$ is a rigid body transformation computed via feature-based image alignment.} This process is discussed further in Section \ref{sec:implementation}.

After associating new measurements, the observed intensities $I_k^{(i)}$ are fused into the map belief. Since the voxels are assumed independent \eqref{eq:independent_assumption}, each voxel is updated separately with
\begin{align} \label{eq:belief_update}
    b_k^{(i)} = \tau(b_{k-1}^{(i)},I_{k+1}^{(i)})
\end{align}
where $\tau$ is the belief update applied to each voxel. At this step, we check the boundary condition and only update the voxels in the current view. We provide additional details \rev{about the estimator} in Section \ref{sec:implementation}.


\subsection{Predictive Perception} \label{sec:predictivePerception}
In this section, we discuss how we can predict the performance of the perception system based on the map belief. This predicted performance will be used to evaluate future plans (hence, perception-aware decision making).

\subsubsection{Predictive Perception Problem}
As discussed in the previous sections, the data history is compressed into the map belief $\mathcal{B}_{k}$. Thus, the predictive perception problem is stated as follows: Given an initial map belief $\mathcal{B}_{k}$ and a future trajectory $\textbf{x}_{k:t}$, predict the measurements $\textbf{z}_{k:t}$ and the associated perception performance $\zeta$. Here, the performance of \ac{VO} is considered for $\zeta$ given image measurements captured by the camera system.

\subsubsection{Perception Performance Prediction}
\begin{figure}[!ht] 
	\centering
	\includegraphics[width=0.8\columnwidth]{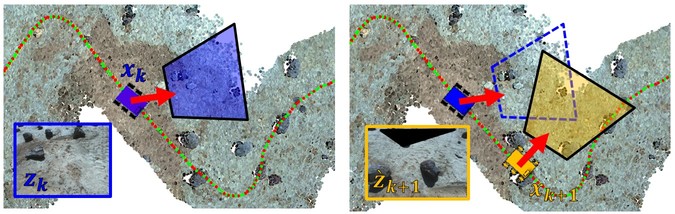}
	\caption{The most likely image sequence is predicted based on the current map belief and future trajectory. The perception performance is evaluated on the resulting image sequence.}
	\label{fig:prediction_illustration}
\end{figure}

Since the measurements are corrupted by noise, the best one can do is find a distribution over all possible future measurements given the current map belief and future trajectory $ p(\textbf{z}_{t} | \mathcal{M}_{k}, \textbf{x}_{t})$. To simplify the problem, we aim to predict only the most likely future image measurements,
\begin{equation}
\grave{\textbf{z}}_{t} = \arg\max_{\textbf{z}_{t}} p(\textbf{z}_{t}|\mathcal{M}_{k},\textbf{x}_{t}).
\end{equation}
This can be done by back-projection using \eqref{eq:camera_projection_model}. Note to recover the visual information (i.e. pixel intensity $I_k^{(i)}$) at the pixel coordinate $\mathbf{u}_k^{(i)}$, 2D interpolation is performed using neighboring pixels on the image plane,
\begin{align}
    I_k^{(i)} = \mathcal{I}(\grave{\mathbf{z}}_k,\mathbf{u}_k^{(i)}),
\end{align}
where $\mathcal{I}$ is an arbitrary interpolation function that operates on the neighborhood of $\mathbf{u}_k^{(i)}$. This process is illustrated in \autoref{fig:prediction_illustration}. 

The perception performance, $\zeta$, is predicted based on the most-likely image sequence, $ \grave{\textbf{z}}_{k:t} $, with respect to the function, $ J $, that defines the planning objective,
\begin{align}
\zeta = J(\grave{\textbf{z}}_{k:t}),
\end{align}
where the computation of $J$ is discussed further in the following sections. Note the predicted measurements are based on the map belief and future trajectory; thus, the performance can also be written as
\begin{align}
\zeta = J(\textbf{x}_{k:t}, \mathcal{B}_{k}).
\end{align}


\subsection{Planning} \label{sec:planning}
Planning under uncertainty typically consists of two primary components: 1) a metric that defines the uncertainty or undesirable behavior of the system, and 2) an optimization procedure for minimizing the defined metric.

\subsubsection{Planning Objective}
In this work, the performance of \ac{VO} is considered as the primary metric for planning. From the previous sections, we showed how to construct the map belief $\mathcal{B}_k$ and predict the most likely future measurements $ \{\grave{\textbf{z}}_{t} \} $ for all $ t>k $. For a specified \ac{VO} algorithm, the performance is evaluated by applying \ac{VO} on the predicted measurements (i.e. synthetic images). 

In other words, the \ac{VO} algorithm is applied to evaluate a potential action by using a set of synthetic images for future time-steps $i$ and $i+1$ to predict the motion and corresponding performance:
\begin{align}
\mathrm{VO}: (\grave{\textbf{z}}_i, \grave{\textbf{z}}_{i+1}) \rightarrow (\Delta \grave{\textbf{x}}_{i,i+1}, \Omega_{i,i+1}),~~~~i \in \{k,\cdots,t\},
\end{align}
where $ \Delta \grave{\textbf{x}}_{i,i+1} $ is the displacement between time-steps $i$ and $i+1$ computed by \ac{VO} and $\Omega_{i,i+1}$ is the performance indicator for \ac{VO} between time-steps $i$ and $i+1$. Note the performance indicator is designed based on the \ac{VO} algorithm; thus, the performance can be evaluated using metrics such as the number of features, uncertainty in pose, or position error to name a few.

In the remainder of the paper, we aim to find the trajectory that minimizes the position error for a given \ac{VO} algorithm; therefore, we choose the following cost function as the planning objective:
\begin{align}\label{eq:pose_drift}
J(\textbf{x}_{k:t}, \mathcal{B}_{k}) := \sum_{i=k}^{t-1} \gamma^{i-k} || \Delta \grave{\textbf{x}}_{i,i+1} - \Delta \textbf{x}_{i,i+1} ||,
\end{align}
where $\Delta \textbf{x}_{i,i+1}$ is the true simulated displacement between future states $\textbf{x}_{i}$ and $\textbf{x}_{i+1}$, $\Delta \grave{\textbf{x}}_{i,i+1}$ is the estimated displacement between synthetic images $\grave{\textbf{z}}_i$ and $\grave{\textbf{z}}_{i+1}$, and $\gamma$ is a discount factor used to weight nearby observations more than father ones.

Using the provided metric, the observation planning problem is formalized as finding the trajectory that minimizes the planning objective:
\begin{align}\label{eq:generalOpt}
\textbf{x}^{*}_{k:t}=\arg\min_{\textbf{x}_{k:t}\in X_{k:t}} J(\textbf{x}_{k:t}, \mathcal{B}_{k}),
\end{align}
where $ X_{k:t} $ is the set of all possible extended state trajectories.

\subsubsection{Search Method}\label{sec:VOSAP}
Since the problem in \eqref{eq:generalOpt} is computationally intractable, sampling-based methods are employed to reduce the uncountably infinite space of trajectories $X_{k:t}$. In particular, we modify the sampling-based methods Probabilistic Roadmap (PRM) and Optimal Rapidly-exploring Randomized Trees (RRT*) to adapt them for observation planning. We refer to this planner as the VO-aware Sampling-based Planner (VOSAP).

This is achieved by initially creating a PRM in the body configuration space $X^{body}$. The PRM captures the connectivity of free space and provides a set of candidate trajectories in the body configuration space leading to the goal. The unobserved portion of the map is assumed free for the purpose of creating the PRM. Given the velocity profile on edges and the path on the PRM, we define $\pi^{body}(t)$ as a continuous-time function that can be queried to generate a body pose at any given time along the path.

Using the PRM, observation planning is performed by determining \textit{when} and \textit{how} observations should be made along the path. This problem is cast as an optimization problem where we minimize the cost function provided in \eqref{eq:pose_drift}. The optimization problem is further adapted to VO-aware planning by adding the appropriate constraints:
\begin{align} \label{eq:minimization}
\{\tau_{0:n}, \textbf{x}_{\tau_{0:n}}^{mast}\}
=& \arg\min J(\textbf{x}_{\tau_{0:n}}^{body}, \textbf{x}_{\tau_{0:n}}^{mast}, \mathcal{B}_{k}) \\
\mbox{s.t.} \quad
& k = \tau_0 \leq \cdots \leq \tau_i \leq \cdots \leq \tau_n = t  \label{eq:cons_timeOrder} \\
& \textbf{x}^{body}_{\tau_i} = \pi^{body}(\tau_{i})  \label{eq:cons_timeSampling} \\
& \textbf{x}^{mast}_{\tau_i} \in X^{mast}(\textbf{x}^{body}_{\tau_i}) \label{eq:cons_observable} \\
& d(\textbf{x}^{mast}_{\tau_i}, \textbf{x}^{mast}_{\tau_{i+1}}) < D_{overlap} \label{eq:cons_overlap}.
\end{align}
A detailed description of the constraints are provided in the following section. 

\rev{In this formulation, the problem is extended to the temporal-spatial space $ \mathcal S = \{\mathcal{T}, X^{mast}\} $ where $\mathcal{T}$ is the domain of $\pi^{body}(t)$ and $X^{mast}$ is the set of all possible mast configurations.} To solve this optimization problem, we propose a spatio-temporal RRT* algorithm presented in Algorithm \ref{alg:RRT}. The algorithm works by incrementally building a tree $\mathcal{G}$ by sampling new points in the state space and connecting them to the existing nodes in the tree. At each iteration, the algorithm samples a new time from the domain of the PRM and a mast configuration from the corresponding configuration space and adds the node to the tree so the cost is minimized at this step. If all constraints are not satisfied, the node is not added to the tree. 

The functions \verb|Predecessors()| and \verb|Successors()| are used for searching the nodes in the tree that satisfy the temporal conditions. \verb|Predecessors()| finds nodes that have earlier time stamps than the query node, and \verb|Successors()| finds nodes with later time stamps. From the generated tree, we can query the best sequence of observations to maximize the perception performance. \rev{This search method requires computing the cost (i.e., \ac{VO}) on a large set of synthetic images, which might result in undesirable computational requirements. Thus, alternative performance metrics are proposed and compared in later sections with the aim of reducing the overall runtime of the algorithm.}


\begin{algorithm}[t]
	\caption{Spatio-temporal RRT*}
	\label{alg:RRT}
	\begin{algorithmic}
		\State $ \mathcal{G} \leftarrow \{ \textbf{x}_{init} \}  $
		\For{$ i  = 1 \to N $}
		\State $ \tau_{new} \leftarrow \mbox{SampleTime}() $
		\State $ \textbf{x}_{new}^{body} \leftarrow \pi^{body}(\tau_{new}) $
		\State $ \textbf{x}_{new}^{mast} \leftarrow \mbox{SampleMastTarget}() $
		\State $ \textbf{x}_{new} \leftarrow \{\textbf{x}_{new}^{body},\textbf{x}_{new}^{mast}\} $
		\State $ c_{min} \leftarrow \infty $	
		\For{$ \textbf{x}_{pred} \in \mbox{Predecessors}(\mathcal{G}, \tau_{new}) $}
		\If{$ \mbox{SatisfyConstraints}( \textbf{x}_{pred}, \textbf{x}_{new} ) $ and \\
			\hspace{2cm} $ \mbox{Cost}( \textbf{x}_{pred}, \textbf{x}_{new} ) < c_{min} $ }
		\State $ \textbf{x}_{min} \leftarrow \textbf{x}_{pred} $
		\State $ c_{min} \leftarrow \mbox{Cost}( \textbf{x}_{pred}, \textbf{x}_{new} ) $
		\EndIf				
		\EndFor  \Comment Find minimum-cost predecessor
		\If{$ c_{min} < \infty $}
		\State $ \mbox{Connect}(\textbf{x}_{min}, \textbf{x}_{new}) $
		\For{$ \textbf{x}_{succ} \in \mbox{Successors}(\mathcal{G}, \tau_{new}) $}
		\State $ \textbf{x}_{parent} \leftarrow \mbox{Parent}(\textbf{x}_{succ} ) $
		\If{$ \mbox{SatisfyConstraints}( \textbf{x}_{new}, \textbf{x}_{succ} ) $ and \\
			\hspace{2cm}  $ \mbox{Cost}( \textbf{x}_{new}, \textbf{x}_{succ} ) < \mbox{Cost}( \textbf{x}_{parent}, \textbf{x}_{succ} ) $}
		\State $ \mbox{Reconnect}(\textbf{x}_{new}, \textbf{x}_{succ}) $
		\EndIf				
		\EndFor
		\EndIf  \Comment Add node and rewire tree
		\EndFor
	\end{algorithmic}
\end{algorithm}

\subsubsection{Planning Constraints}
The planning constraints used in the optimization procedure are briefly described in this section.
Let $ \textbf{x}_{i}^{mast} $ denote the mast configuration when taking the $ i $-th image. Here, the mast configuration is represented as a 2D point on the ground plane referred to as the mast target. The mast target is the intersection of the camera focal line and the ground plane.

\paragraph{Time Constraints}
Equation \eqref{eq:cons_timeOrder} restricts the algorithm to search causal nodes from the tree. In other words, future locations must have greater associated time stamps. Further, the search space can be reduced by employing another constraint:
\begin{align}
| \tau_{new} - \tau_{i} | < T_{search},
\end{align}
where $ T_{search} $ is a constant to limit the size of the search window in time space.

\paragraph{Field-of-view Constraint}
In \eqref{eq:cons_observable}, a constraints is applied in the planning framework due to the \ac{FOV} of the sensor. Due to the body occlusion and kinematic constraints of the mast manipulator, the mast target is limited to a ring around the rover. The applied constraint is given by
\begin{align}	
D_{near} \leq || \textbf{x}^{mast}_{i} - \textbf{x}^{body}_{i} || \leq D_{far},
\end{align}
where $ D_{near} $ and $ D_{far} $ are the inner and outer radius of the \ac{FOV} ring (around the robot) on the ground.

\paragraph{Image-overlapping Constraint}
In order for \ac{VO} to operate successfully, the constraint in \eqref{eq:cons_overlap} is applied to ensure images overlap. Since the mast target is represented as a point on the ground, the amount of overlap between images can be conservatively approximated using the Euclidean distance:
\begin{align}
d(\textbf{x}^{mast}_{i}, \textbf{x}^{mast}_{i+1})=||\textbf{x}^{mast}_{i} - \textbf{x}^{mast}_{i+1}|| < D_{overlap}.
\end{align}

\subsubsection{Receding Horizon Planning}
Due to discrepancies between computational and real-world models, hardware restrictions of the perception systems (e.g, camera resolution), and other non-idealities, it is not possible plan for lengthy horizons. To cope with this challenge, the VOSAP algorithm is wrapped in an outer replanning loop using a \ac{RHC} scheme for dynamic replanning. 

Recently, RHC methods have been extended to systems with stochastic dynamics and noisy perception  \cite{agha2018slap,Erez2010,Platt10,Chakrav11-IRHC,He11JAIR,Toit10}. 
In the RHC setting, optimization is carried out only within a limited horizon and returns a sequence of optimal actions within the horizon. The system only takes the first (or the first few) action(s) of this optimal sequence. Then, it obtains new measurements, updates its belief over the robot location and the environment map, replans from the new belief, and repeats this process until it reaches the goal point. The \ac{RHC} approach allows the algorithm to plan paths under a limited sensing field of view.


\section{Implementation}
\label{sec:implementation}
In general, each component of the algorithm (e.g. mapping, prediction, and planning) is independent and can be interchanged with variations. Therefore, in this section, we present further details for our implementation to allow the reader to more easily replicate our method.

\subsection{Map Initialization}
In order to initialize the map, multiple modes are used for image registration. If information is available from external sources (e.g. helicopter, orbiter), the map is initialized using feature-based image alignment. Otherwise, if no external information is available, multiple images are captured and projected to the map using the extended state of the rover. In this work, we assume the ground is planar, and therefore, the projection model can be represented as a homography given by
\begin{align} \label{eq:voxel_to_pixel_implementation}
    \mathbf{u}^{(i)}_k = {}^k\mathbf{H}_c {} \mathbf{q}^{(i)}_k
\end{align}
where $\mathbf{q}_k^{(i)} = [x_k^{(i)},y_k^{(i)},1]^\top$ is the projective coordinates of the $i$-th voxel in the plane (e.g. $z=0$) and ${}^k\mathbf{H}_c$ is the homography that projects a voxel in the camera reference frame to a point in the image plane at time step $k$. Note the inverse projection is valid for this representation of the map, and the homography ${}^k\mathbf{H}_c$ can be extracted from the extended state of the rover and the camera projection model using the ground as the model plane \cite{zzhang2000}. 

\begin{figure}[h] 
	\centering
	\includegraphics[width=0.6\columnwidth]{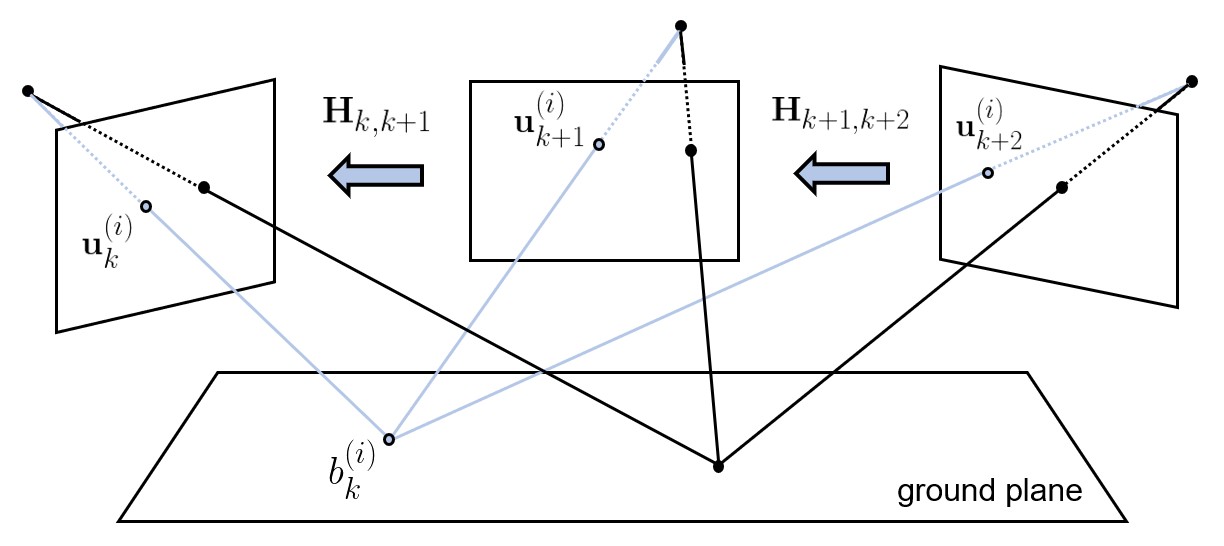}
	\caption{Illustration of feature-based image alignment process used for associating image measurements to the map belief.}
	\label{fig:image_alignment}
\end{figure}

\rev{After initializing the map, the algorithm would ideally only require images captured for \ac{VO} to update the map; however, additional images must be captured since the gaze of the cameras are bias towards regions containing the most visual texture. Thus, images captured for \ac{VO} do not always cover the space visible from the viewpoint of the camera at future body poses. If this occurs, the planner will never be capable of finding a feasible trajectory for the full length of the specified horizon. To account for this fact, if the planner fails to plan a trajectory for the full horizon length, multiple images are captured using the same sequence of images as for initialization. The captured images are then used to update the map using feature-based image alignment.}

\subsection{Feature-Based Image Alignment}
For updating the map, feature-based image alignment is performed to compute the transformation between image measurements. In contrast to direct methods, a feature-based approach was chosen for this application to support matching images with large disparity. Using a planar representation of the map, the relationship between the $i$-th and $j$-th image is given by
\begin{align} \label{eq:image_alignment}
    {}^i\mathbf{H}_{j} = \prod_{k=i}^{j} {}^{k-1}\mathbf{H}_{k}
\end{align}
where the homography between each image is computed using normalized \ac{DLT} in a RANSAC scheme as described in \cite{Hartley2000}. This process is visualized in \autoref{fig:image_alignment}. In this step, we use ORB features \cite{Rublee11_ORB}, which provides a sufficient level of efficiency and accuracy. 

From \autoref{eq:voxel_to_pixel_implementation} and \autoref{eq:image_alignment}, the relationship between the $k$-th image and the map belief can be computed as
\begin{align} \label{eq:image_alignment_implementation}
    \mathbf{q}^{(i)}_k = \left( {}^m\mathbf{H}_{0} \prod_{j=1}^{k} {}^{j-1}\mathbf{H}_{j} \right) \mathbf{u}^{(i)}_k,
\end{align}
where $\mathbf{H}_{m,0}$ is the homography used for initializing the map. In the case image alignment fails (e.g. feature-poor terrain), the map is reinitialized using the previously described procedure.

\subsection{Perception Performance}
While the proposed approach is agnostic to the specific \ac{VO} algorithm used for computing the cost, we assume the ground is planar. As a result, methods for \ac{VO} that are unstable for coplanar points (e.g. fundamental matrix) are not valid for this representation. We exploit this fact and compute the displacement from the euclidean homography,
\begin{align} \label{eq:displacement_homography}
    \Delta \grave{\mathbf{x}}_{k,k+1} = || \grave{\mathbf{g}}_{k+1} - \grave{\mathbf{g}}_{k} ||
\end{align}
where $\grave{\mathbf{g}}_{k}$ and $\grave{\mathbf{g}}_{k+1}$ are the positions of the rover extracted from the euclidean homography at time step $k$ and $k+1$, respectively. Note the displacement is estimated in this way only for synthetic images. 


\begin{figure}[h]
	\centering
	\includegraphics[width=1\columnwidth]{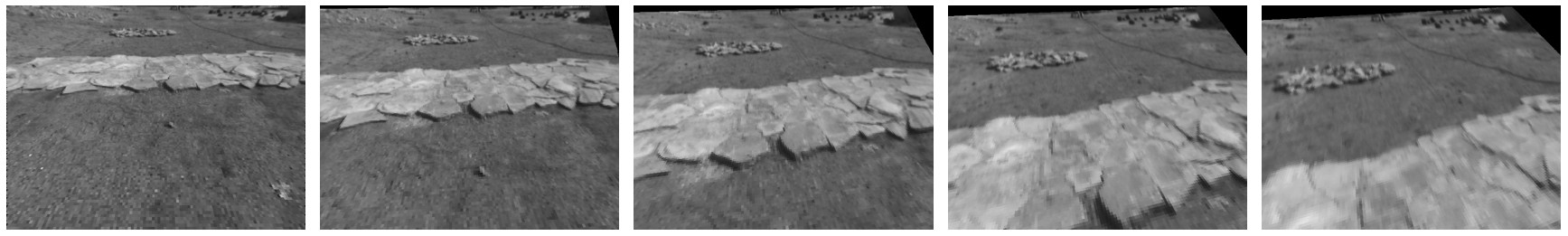}
	\caption{Image sequence $\grave{\mathbf{z}}_{k:t}$ predicted from a single view. The leftmost image is the original measurement used for constructing the map. The right images are the most likely images for a potential trajectory $\mathbf{x}_{k:t}$.}
	\label{fig:example_image_sequence}
\end{figure}

\subsection{Kalman Update}
In this work, we use a Kalman filter to update the voxel beliefs at each time step. Since the map is assumed stationary, the process model is the identity; thus, the belief at each voxel can be updated as
\begin{align} \label{eq:kalman_update}
    P_{k|k} = \big( P_{k|k-1}^{-1} + &R^{-1} \big)^{-1} \\
    I_{k|k} = P_{k|k} \Big( P_{k|k-1}^{-1} I_{k|k-1} + &R^{-1}\mathbf{z}_{k}(\mathbf{u}_k) \Big)
\end{align}
where $I_{k|k-1}$ and $P_{k|k-1}$ are the mean and variance of the visual information at the $i$-th voxel given the belief at time step $k-1$. The variance $P_{k|k-1}$ is computed by adding process noise $Q$. The observation noise $R$ and process noise $Q$ are assumed to be distributed normally.


\section{Field Experiments} \label{sec:field_experiments}
\rev{As ambitious future missions rely on increased autonomy, manual mast pointing is not acceptable due to a number of factors such as strict operational timelines and communication constraints (e.g., signal delay, blockage, bandwidth, and availability of the \ac{DSN}). For example, Mars 2020 aims to operate on a 5-hour operational timeline, which is not possible with the current mode of operations (e.g., manual commands). As a result, future missions must rely on either automated mast pointing or a static mast for perception. Thus, the proposed method for automated mast pointing is validated using a static mast as the baseline. In previous work, the method was evaluated using a high-fidelity physics-based simulator, \ac{ROAMS}, \cite{roams2003}. In this work, we focus on validating the proposed method on actual hardware in a Mars-analogue environment.}


\subsection{Experimental Setup} \label{sec:experimental_setup}

\subsubsection{Software Framework}
The system is composed of a primary computer for high-level navigation and a secondary computer for real-time tasks. The programming language used for all software is C++ and was developed in Linux using the CLARAty framework \cite{Nesnas06_claraty}, which is a framework for generic and reusable software that can be adapted for different research platforms. For interfacing between the primary and secondary computers (as well as CLARAty modules), we use \ac{ROS}.

\subsubsection{Hardware Platform}
The hardware platform used for the experiments is the Athena rover developed at \ac{JPL} \cite{Biesiadecki01_Athena}. The platform is designed for testing Mars rover capabilities on Earth and is comparable in size to \ac{MER} \cite{Maki2003}. The platform has been updated from the original design of the system, and now, an Nvidia TX1 is used as the main computer for navigation and planning, and an Nvidia TK1 is used as the secondary computer for interfacing hardware. The navigation system is primarily vision-based using a stereo-camera pair consisting of two PointGrey Flea2 cameras rigidly mounted on a movable mast with pan-tilt capabilities. The mast is at a height of 1.4\,m, and the baseline of the stereo-camera pair is 0.30\,m. The images are captured at a resolution of 640x480 from 82x66 degree field-of-view lenses. The pose of the mast relative to the body is obtained via encoders on the motors used for driving the mast. During the experiments, the rover drives at approximately 0.042 meters per second.

\begin{figure}[!ht]
	\centering
	\includegraphics[width=0.45\columnwidth]{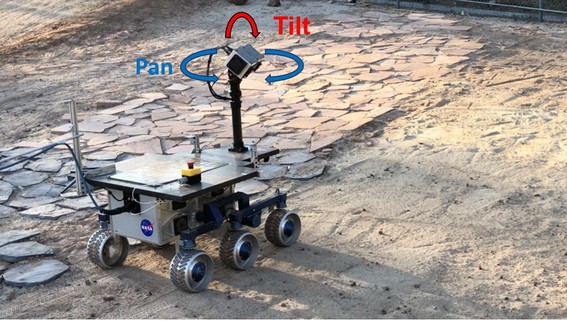}
	\caption{Athena rover in the Mars Yard during field experiments.}
	\label{fig:athena_diagram}
\end{figure}

\subsubsection{Environment Description}
To \rev{validate} the proposed method in a realistic environment, the experiments are performed at the \ac{JPL} Mars Yard. The facility is a large outdoor environment 21x22\,m in size designed to mimic the Martian landscape. This environment is currently used for research and prototyping of robotic platforms for space flight projects. The terrain is designed to match images from current Mars missions and consists of sand and rocks of various shapes and sizes providing a variety of potential arrangements for the landscape. The experiments are performed in regions of the Mars Yard containing a mixture of both feature-poor and feature-rich terrains.

\subsubsection{Mars Yard versus Mars}
\label{sec:marsyard_vs_mars}
While the Mars Yard provides a environment comparable to the surface of Mars, the terrain in the Mars Yard is not as difficult to navigate as the terrain on Mars in regards to vision-based navigation systems. For example, consider the smooth sand in \autoref{fig:illustration} on the surface of Mars in comparison to the sand in \autoref{fig:athena_diagram} at the Mars Yard. 
\rev{In the Mars Yard, the sand provides a reasonable amount of texture in comparison to Mars, and as a result, the relative performance over the baseline is expected to improve even further if applied to rovers on Mars where texture-limited surfaces are more prevalent and severe.}
To account for this fact, the sand was raked prior to each experiment to replicate the smooth sand regions on Mars.


\subsection{\rev{Comparison of Passive and Active Perception}} \label{sec:evaluation_active_passive}
\rev{To validate the proposed algorithm on the described system, a set of experiments are performed using both the optimal mast trajectory estimated using VOSAP (active perception) and a fixed-mast trajectory (passive perception).}

\subsubsection{Description of Experiments}
For the fixed-mast trajectory, the camera is tilted downward at an angle of $30$ degrees for the duration of each experiment. For VOSAP, we limit the range of mast motion to a close proximity around the rover. The tilt angle is limited between $30$ to $45$ degrees downward and the pan angle is limited between $-90$ to $90$ degrees left to right. In order to provide an unbiased comparison, the rover acquires images for both the passive and active mast configurations \rev{during each experiment}. This is achieved by keeping the rover stationary during image acquisition and moving the mast to the corresponding passive and active configurations at each step along the trajectory. Note the initial mast pose is identical for both the passive and active mast configurations.

In this work, we perform 10 different field experiments during different times of day. Experiments 1-5 were performed in the evening during low solar elevation, and experiments 6-10 were performed in the afternoon during high solar elevation. \rev{For each experiment, the rover was not given a prior map. Thus, three images were captured to initialize the map using a tilt angle of 30 degrees (downward) with pan angles -45, 0, and 45 degrees. The rover updates the map online using the previously described methods, and if the map is degenerate (i.e., the map does not cover future viewpoints), additional images are captured using the initialization sequence to update the map as discussed in Section \ref{sec:implementation}}. Additionally, we rely on the multi-step RHC algorithm as the outer loop of VOSAP. Initially, the rover is given a goal, and a continuous body trajectory $\pi^{body}(t)$ is computed for reaching the goal. At each step, the map is updated, and the planner computes the optimal mast trajectory with a horizon of 3 meters. Then, the rover executes the first command from the planned sequence of optimal mast configurations and repeats this process until reaching the goal.

\begin{table}[!t]
\caption{Error of VO for field experiments} \label{tab:experimental_results_numbers}
    \small
	\centering
	\begin{tabular}{p{2.5cm}p{0.9cm}p{0.9cm}p{0.9cm}p{0.9cm}p{0.9cm}p{0.9cm}p{0.9cm}p{0.9cm}p{0.9cm}p{0.9cm}}
		\toprule
		& 1 & 2 & 3 & 4 & 5 & 6 & 7 & 8 & 9 & 10 \\
		\midrule
		~~~Passive (m) & 0.224 & 0.988 & 0.364 & 1.579 & 1.257 & Failed & 0.148 & 0.583 & 0.266 & Failed  \\
		~~~Active (m) & 0.131 & 0.649 & 0.240 & 0.488 & 0.733 & 0.057 & 0.065 & 0.101 &  0.211 & 0.552 \\
		~~~Improvement & 42.4\% & 34.4\% & 34.0\% & 69.1\% & 41.7\% & -------- & 55.9\% & 82.7\% & 20.8\% & -------- \\
		\bottomrule
	\end{tabular}
\end{table}

\begin{figure*}[!t]
\centerline 
{
    \hspace{-1\baselineskip}
	\subfigure[Trajectory (Exp 1)]
	{
		\includegraphics[width=\myvala\columnwidth]{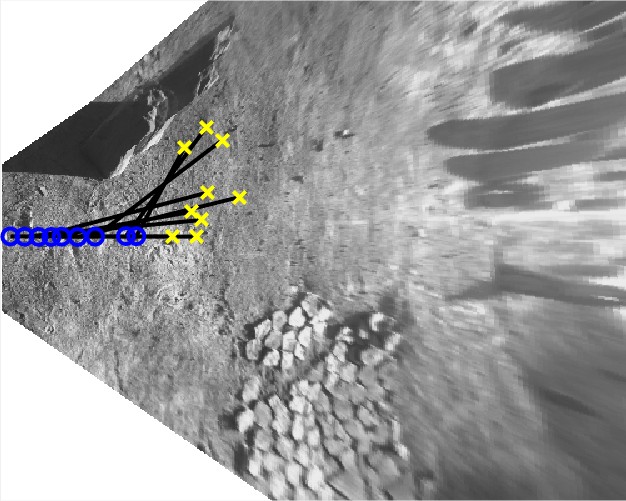}
		\label{fig:run1_traj}
	}
	\hspace{-1\baselineskip}
	\subfigure[Trajectory (Exp 2)]
	{
		\includegraphics[width=\myvala\columnwidth]{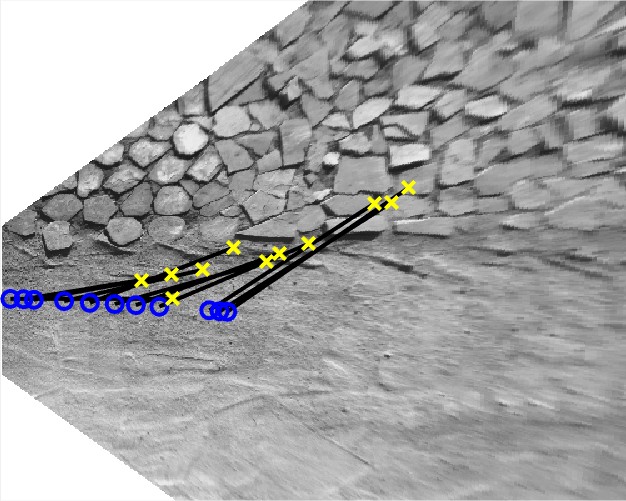}
		\label{fig:run2_traj}
	}
	\hspace{-1\baselineskip}
	\subfigure[Trajectory (Exp 3)]
	{
		\includegraphics[width=\myvala\columnwidth]{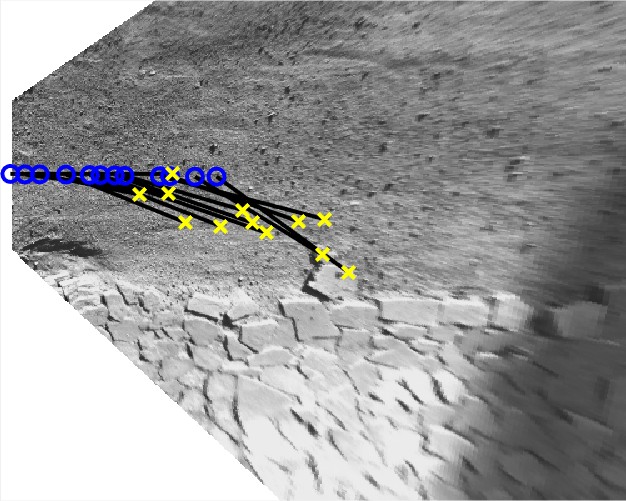}
		\label{fig:run3_traj}
	}
	\hspace{-1\baselineskip}
	\subfigure[Trajectory (Exp 4)]
	{
		\includegraphics[width=\myvala\columnwidth]{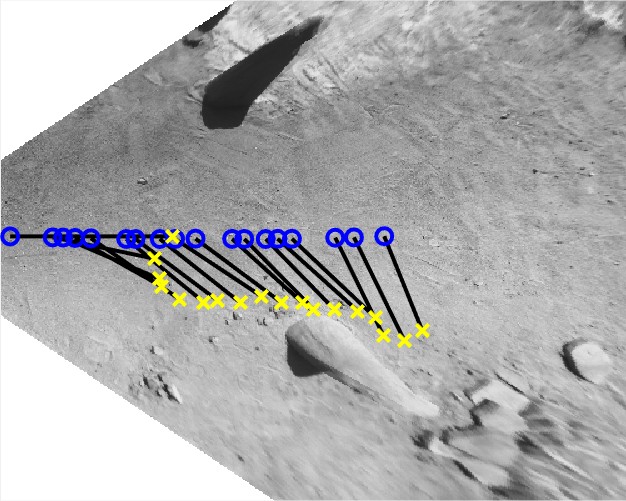}
		\label{fig:run4_traj}
	}
	\hspace{-1\baselineskip}
	\subfigure[Trajectory (Exp 5)]
	{
		\includegraphics[width=\myvala\columnwidth]{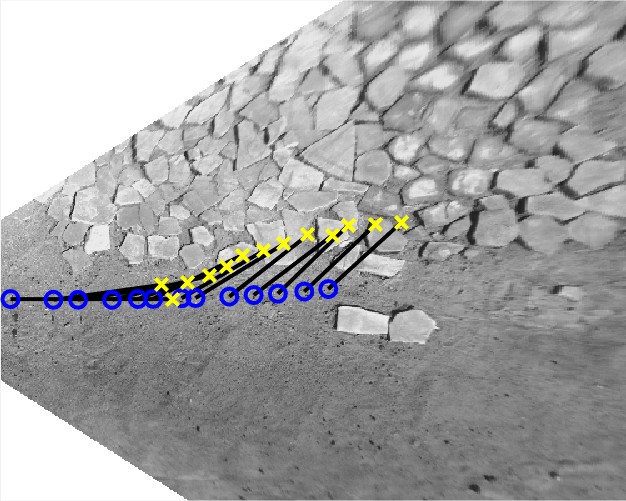}
		\label{fig:run5_traj}
	}
}
\centerline  
{
    \hspace{-1\baselineskip}
	\subfigure[Trajectory (Exp 6)]
	{
		\includegraphics[width=\myvala\columnwidth]{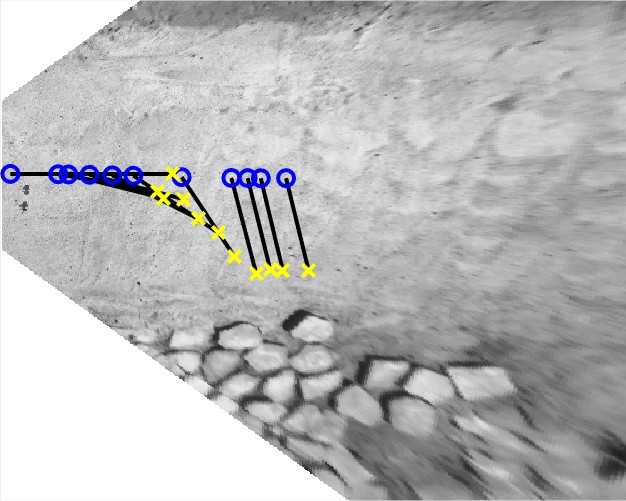}
		\label{fig:run6_traj}
	}
	\hspace{-1\baselineskip}
	\subfigure[Trajectory (Exp 7)]
	{
		\includegraphics[width=\myvala\columnwidth]{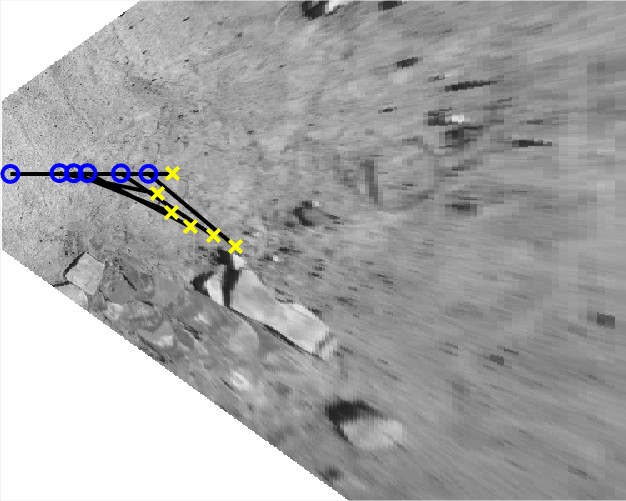}
		\label{fig:run7_traj}
	}
	\hspace{-1\baselineskip}
	\subfigure[Trajectory (Exp 8)]
	{
		\includegraphics[width=\myvala\columnwidth]{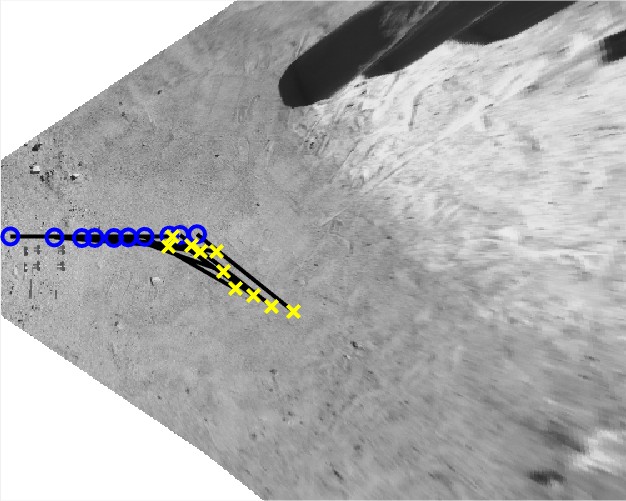}
		\label{fig:run8_traj}
	}
	\hspace{-1\baselineskip}
	\subfigure[Trajectory (Exp 9)]
	{
		\includegraphics[width=\myvala\columnwidth]{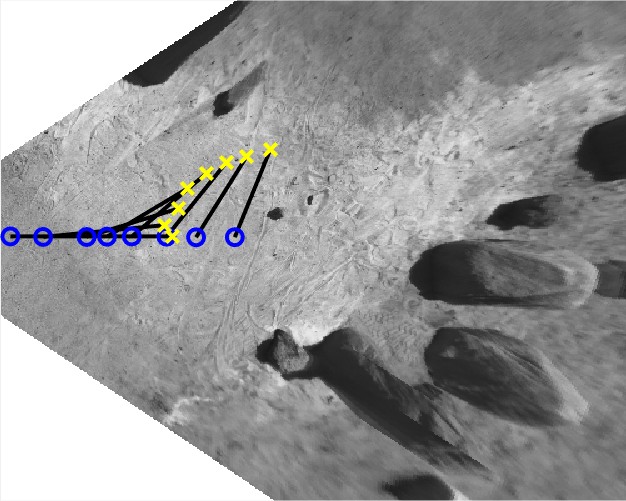}
		\label{fig:run9_traj}
	}
	\hspace{-1\baselineskip}
	\subfigure[Trajectory (Exp 10)]
	{
		\includegraphics[width=\myvala\columnwidth]{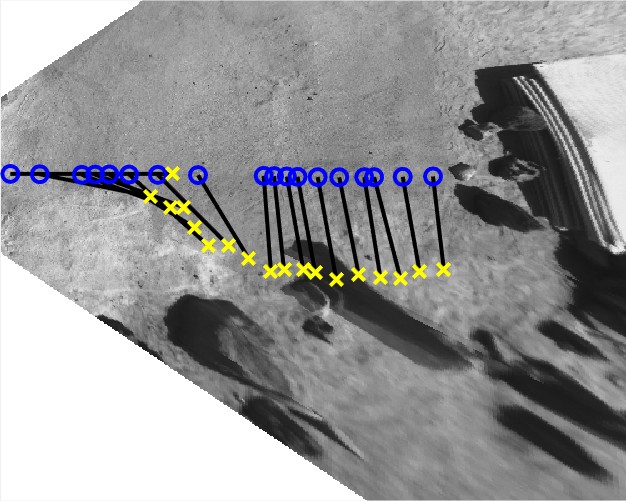}
		\label{fig:run10_traj}
	}
}
\caption{Top-view of the map for each of the field experiments. \rev{The blue circles represent the camera positions, the black lines represent the mast azimuth angle (pan), and the yellow crosses represent the mast elevation angle (tilt) defined by the intersection of the focal line and the ground.}}
\label{fig:experimental_results_trajectory}
\end{figure*}

\subsubsection{Discussion of Results}
The proposed approach is agnostic to the specific \ac{VO} algorithm; therefore, we use a simple feature-based \ac{VO} algorithm using A-KAZE \cite{Alcantarilla11_AKAZE} features for computing \ac{VO} for both the passive and active mast configurations. 
The estimated trajectories are compared against the trajectory computed using a high-frequency \ac{VO} algorithm. 
\rev{The proposed method is validated based on the accuracy of \ac{VO} (i.e., position error) as well as the success rate of \ac{VO}. 
The success (or failure) of each experiment is based on whether \ac{VO} is capable of computing a navigation solution along the entire trajectory.  
The results are reported in \autoref{tab:experimental_results_numbers} showing the \ac{VO} accuracy as well as the improvement in performance of the proposed method (i.e., active) over the baseline (i.e., passive) for each experiment.
The trajectory for each experiment is overlaid on the estimated map and presented \autoref{fig:experimental_results_trajectory}. 
In general, the performance gain is dependent on the environment; thus, the quantitative results do not imply an expected performance gain in general. 
Instead, the algorithm is validated in a Mars-analogue environment, and qualitatively, the results show that the proposed method is successful in steering the mast towards regions with the most visual texture.
As a result, in this particular environment, the proposed method showed a performance gain of at least 20\% in each of the performed experiments with a success rate of 100\% in contrast to a success rate of 80\% for the static mast.
Each of the experiments are discussed in detail in the remainder of this section.}

\begin{figure*}[!t]
\centerline 
{
    \hspace{-1\baselineskip}
	\subfigure[Active (Exp 1)]
	{
		\includegraphics[width=\myvala\columnwidth]{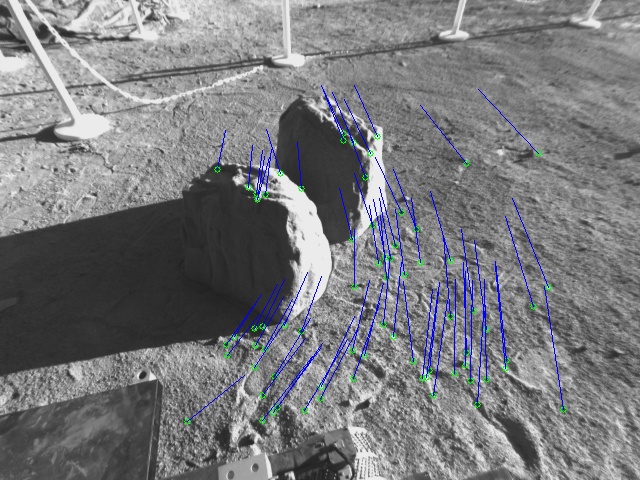}
		\label{fig:run1_tracks_a}
	}
	\hspace{-1\baselineskip}
	\subfigure[Active (Exp 2)]
	{
		\includegraphics[width=\myvala\columnwidth]{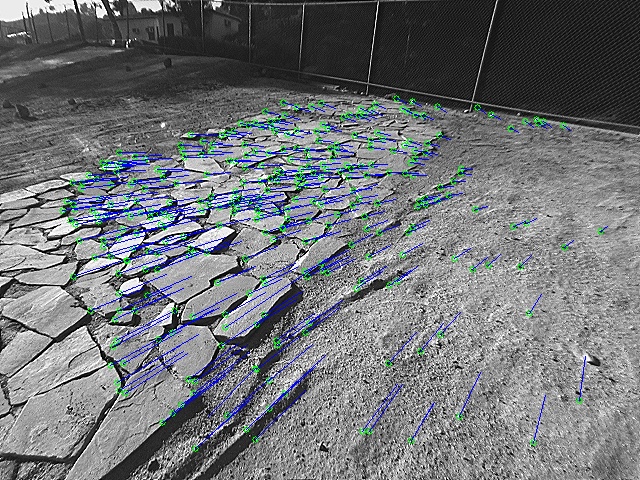}
		\label{fig:run2_tracks_a}
	}
	\hspace{-1\baselineskip}
	\subfigure[Active (Exp 3)]
	{
		\includegraphics[width=\myvala\columnwidth]{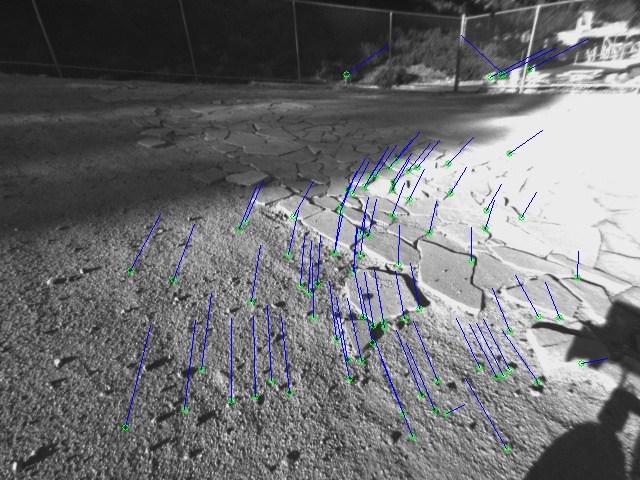}
		\label{fig:run3_tracks_a}
	}
	\hspace{-1\baselineskip}
	\subfigure[Active (Exp 4)]
	{
		\includegraphics[width=\myvala\columnwidth]{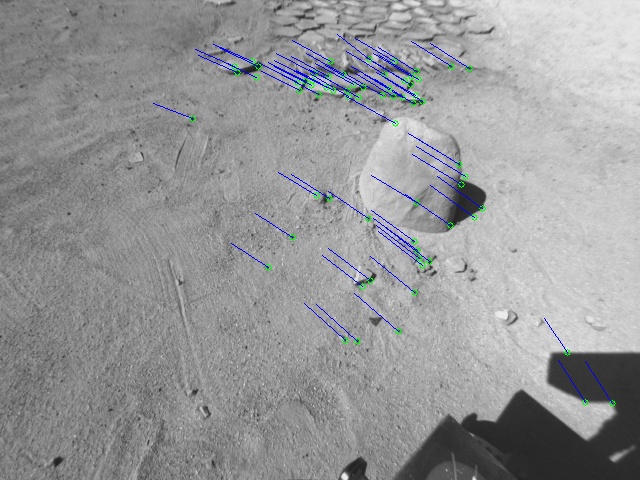}
		\label{fig:run4_tracks_a}
	}
	\hspace{-1\baselineskip}
	\subfigure[Active (Exp 5)]
	{
		\includegraphics[width=\myvala\columnwidth]{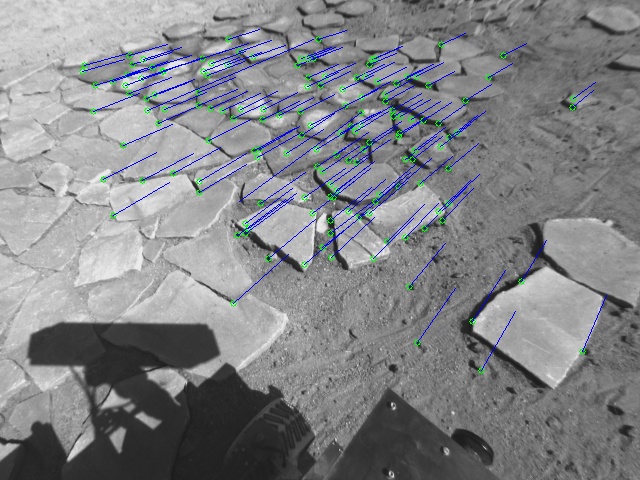}
		\label{fig:run5_tracks_a}
	}
}
\centerline 
{
    \hspace{-1\baselineskip}
	\subfigure[Passive (Exp 1)]
	{
		\includegraphics[width=\myvala\columnwidth]{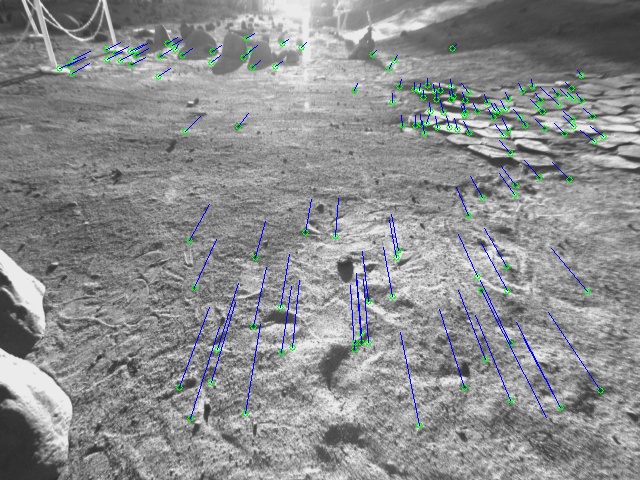}
		\label{fig:run1_tracks_p}
	}
	\hspace{-1\baselineskip}
	\subfigure[Passive (Exp 2)]
	{
		\includegraphics[width=\myvala\columnwidth]{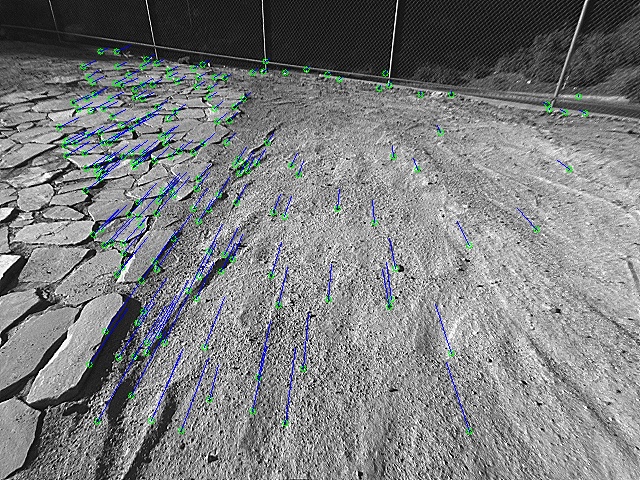}
		\label{fig:run2_tracks_p}
	}
	\hspace{-1\baselineskip}
	\subfigure[Passive (Exp 3)]
	{
		\includegraphics[width=\myvala\columnwidth]{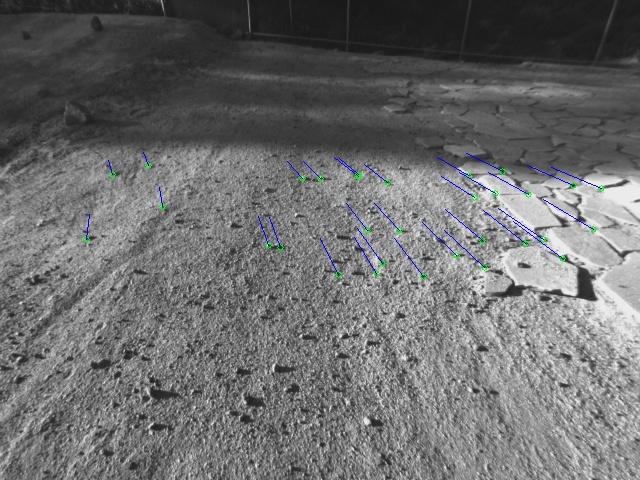}
		\label{fig:run3_tracks_p}
	}
	\hspace{-1\baselineskip}
	\subfigure[Passive (Exp 4)]
	{
		\includegraphics[width=\myvala\columnwidth]{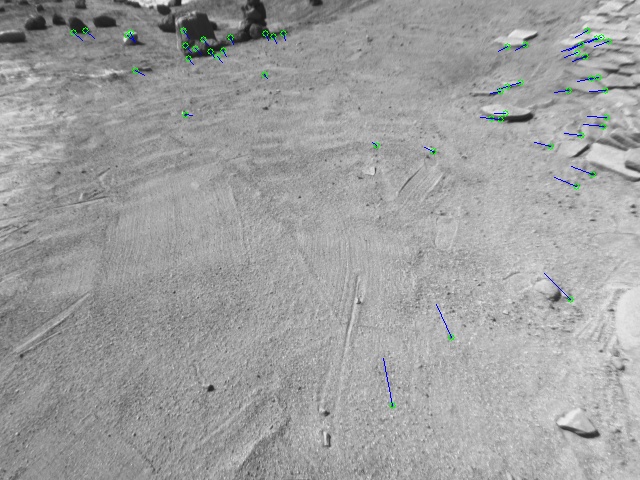}
		\label{fig:run4_tracks_p}
	}
	\hspace{-1\baselineskip}
	\subfigure[Passive (Exp 5)]
	{
		\includegraphics[width=\myvala\columnwidth]{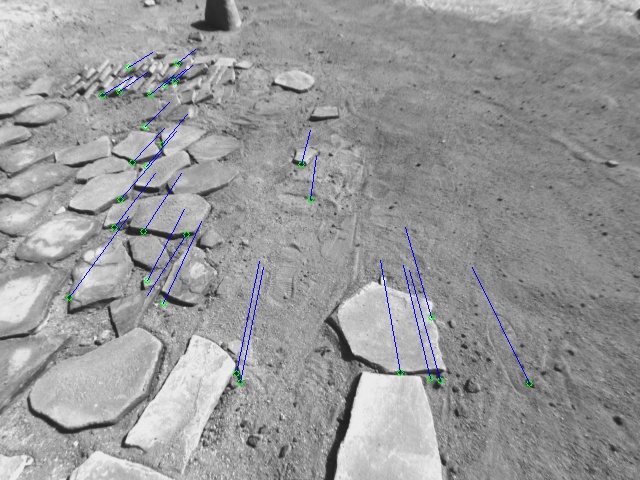}
		\label{fig:run5_tracks_p}
	}
}
\caption{Tracked features using mast trajectory estimated from VOSAP (active) and fixed-mast configuration (passive) for experiments 1-5. Note the tracked features are displayed at the same rover position for each pair of passive and active images.}
\label{fig:experimental_results_features_1_5}
\end{figure*}

The first set of experiments are performed in areas of the Mars Yard containing a modest amount of landmarks. 
In experiments 1 and 4, isolated rocks, bedrock, and sand were evenly distributed throughout the landscape. In experiments 2, 3, and 5, an equal distribution of feature-poor (smooth sand) and feature-rich (bedrock) terrain was arranged on each side of the rover. 
In experiments 1-5, landmarks were visible from the viewpoints of both the passive and active mast configurations.
\rev{Thus, as a result, both passive and active mast configurations were capable of navigating successfully (i.e., without failure) in experiments 1-5. 
However, in this particular environment, the trajectory estimated by VOSAP provided significant improvement in \ac{VO} performance over the static mast in each of the performed experiments. 
This is due to the fact that VOSAP minimizes the \ac{VO} error by aiming the mast towards regions containing a higher quantity and quality of features.}
In experiments 1-5, the experiments were performed in the evening under low solar elevation. This results in an overall lower accuracy for \ac{VO}. \rev{A subset of the images captured for \ac{VO} during experiments 1-5 are presented in \autoref{fig:experimental_results_features_1_5} showing the tracked features for passive and active mast configurations.}

In the remaining experiments 6-10, the rover was set in a more challenging area of the Mars Yard \rev{(e.g. increased sparsity of features).}
In some cases (experiments 6 and 10), the fixed-mast setup failed completely while VOSAP was able to steer the mast toward information-rich areas avoiding \ac{VO} failure. 
In experiment 7, scattered rocks were prevalent in both passive and active mast configurations. 
In both cases, \ac{VO} performed well (in comparison to other experiments); however, VOSAP chose a trajectory that maximized the coverage of rocks in image measurements leading to increased \ac{VO} performance. 
In experiment 8, we expected VOSAP to plan a trajectory facing the upper part of the map towards the more obvious \rev{(i.e., larger)} landmarks; however, VOSAP chose a trajectory facing a set of many isolated rocks leading to improved perception performance. 
In experiment 9, only slight improvement was observed using VOSAP over the fixed-mast trajectory. 
In this case, the result is expected since feature-rich terrain is prevalent for both VOSAP and the fixed-mast setup. 
In experiment 10, isolated landmarks were visible in the upper part of the image \rev{(e.g. near the horizon)} for the passive case, but \ac{VO} failed during the initial portion of the experiment. 
Using VOSAP, \ac{VO} succeeded for the entire trajectory \rev{in experiment 10}.

\begin{figure*}[!t]
\centerline 
{
    \hspace{-1\baselineskip}
	\subfigure[Active (Exp 6)]
	{
		\includegraphics[width=\myvala\columnwidth]{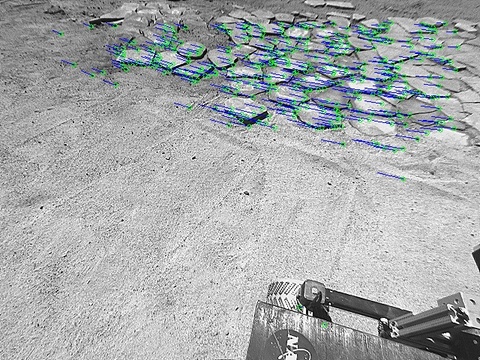}
		\label{fig:run6_tracks_a}
	}
	\hspace{-1\baselineskip}
	\subfigure[Active (Exp 7)]
	{
		\includegraphics[width=\myvala\columnwidth]{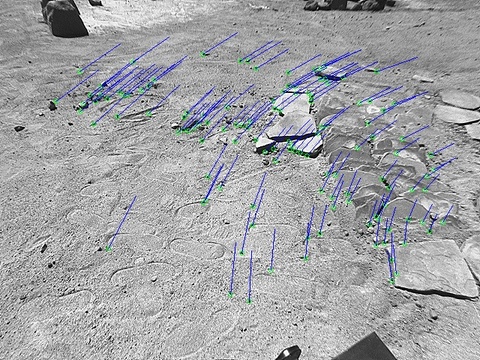}
		\label{fig:run7_tracks_a}
	}
	\hspace{-1\baselineskip}
	\subfigure[Active (Exp 8)]
	{
		\includegraphics[width=\myvala\columnwidth]{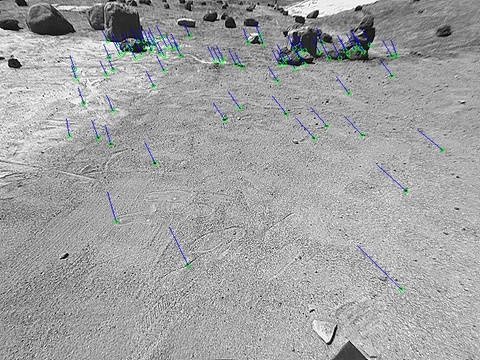}
		\label{fig:run8_tracks_a}
	}
	\hspace{-1\baselineskip}
	\subfigure[Active (Exp 9)]
	{
		\includegraphics[width=\myvala\columnwidth]{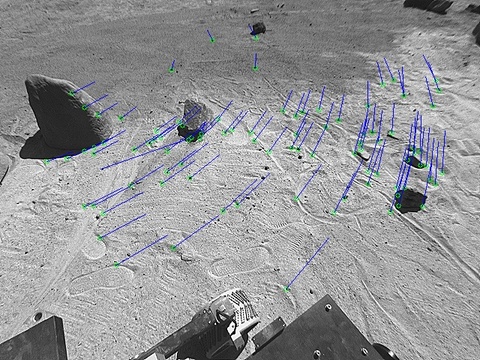}
		\label{fig:run9_tracks_a}
	}
	\hspace{-1\baselineskip}
	\subfigure[Active (Exp 10)]
	{
		\includegraphics[width=\myvala\columnwidth]{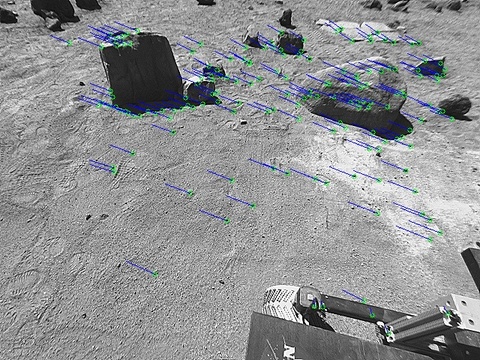}
		\label{fig:run10_tracks_a}
	}
}
\centerline 
{
    \hspace{-1\baselineskip}
	\subfigure[Passive (Exp 6)]
	{
		\includegraphics[width=\myvala\columnwidth]{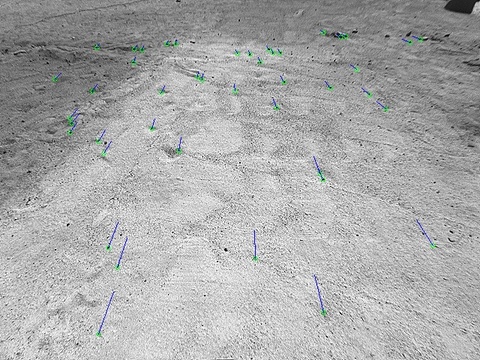}
		\label{fig:run6_tracks_p}
	}
	\hspace{-1\baselineskip}
	\subfigure[Passive (Exp 7)]
	{
		\includegraphics[width=\myvala\columnwidth]{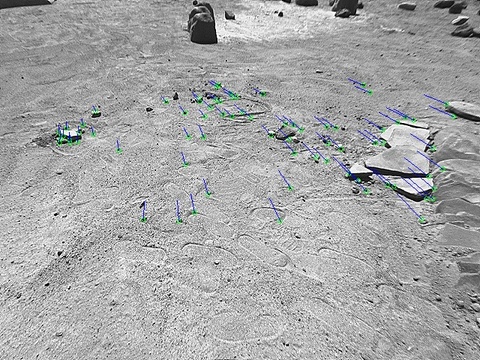}
		\label{fig:run7_tracks_p}
	}
	\hspace{-1\baselineskip}
	\subfigure[Passive (Exp 8)]
	{
		\includegraphics[width=\myvala\columnwidth]{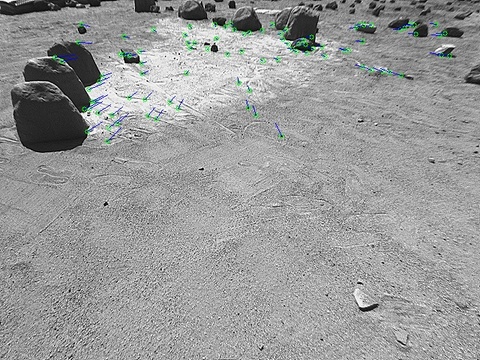}
		\label{fig:run8_tracks_p}
	}
	\hspace{-1\baselineskip}
	\subfigure[Passive (Exp 9)]
	{
		\includegraphics[width=\myvala\columnwidth]{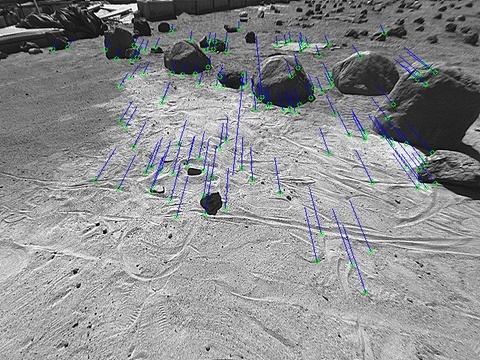}
		\label{fig:run9_tracks_p}
	}
	\hspace{-1\baselineskip}
	\subfigure[Passive (Exp 10)]
	{
		\includegraphics[width=\myvala\columnwidth]{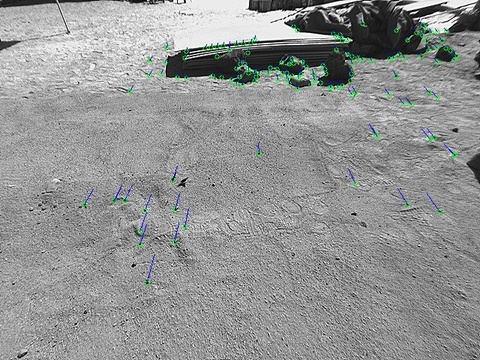}
		\label{fig:run10_tracks_p}
	}
}
\caption{Tracked features using mast trajectory estimated from VOSAP (active) and fixed-mast configuration (passive) for experiments 6-10. Note the tracked features are displayed at the same rover position for each pair of passive and active images.}
\label{fig:experimental_results_features_6_10}
\end{figure*}

\rev{The qualitative results shown for the experiments (\autoref{fig:experimental_results_features_1_5} and \autoref{fig:experimental_results_features_6_10}) show that the presented method does indeed choose trajectories that maximize the quality and quantity of features captured by the visual sensors.
This is particularly clear in \autoref{fig:experimental_results_features_1_5} and even more so in \autoref{fig:experimental_results_features_6_10} as \ac{VO} is capable of tracking significantly more features using the active configurations in contrast to the passive configurations.
Similarly, the steering of the mast towards textured regions is evident in \autoref{fig:experimental_results_trajectory}.
This capability is most clear in experiments 2, 3, 5, and 6 where the ground to one side of the rover is covered in bedrock (feature-rich) and the ground on the opposite side of the rover is covered in sand (feature-poor).
However, in the less obvious cases (e.g., experiment 8), the feature-rich and feature-poor regions are more difficult to identify, but quantitatively, the \ac{VO} accuracy improved in such cases using the active configurations in contrast to the passive configurations.
As mentioned previously, the performance gain is dependent on the environment as well as the configuration of landmarks in the environment, and therefore, the results do not imply a particular performance gain in general.
However, the proposed method is validated as the results shown in \autoref{tab:experimental_results_numbers} show that using VOSAP the accuracy improved significantly over the fixed-mast case in the experiments performed in the Mars Yard.}

\subsection{Comparison of Performance Indicators}
\label{sec:comparison_performance_indicators}
In this work, we directly minimize the localization drift by applying \ac{VO} on synthetic images. While this provides the optimal trajectory in regards to localization drift, this requires computing \ac{VO} on a large set of synthetic images leading to undesirable computational requirements. During the experiments, large correlations were found between localization drift and other performance metrics. Therefore, it is worthwhile to consider alternative indicators of \ac{VO} performance to reduce the overall runtime. This is especially important in space applications where the computation resources are limited due to hardware constraints. We compare the following metrics in the proposed framework:
\begin{itemize}
    \item The displacement error computed on synthetic images (see \autoref{eq:pose_drift}). This is the baseline approach and is described in detail in previous sections. Let this metric be denoted by $J_{D}$.
    \item The number of features in synthetic images. This requires computing only the features in each synthetic image and avoids computing descriptors, matching features, and estimating motion. Let this metric be denoted by $J_{F}$.
    \item The number of features visible from each viewpoint. This is achieved by maintaining a map belief containing only features. This avoids performing operations on synthetic images. Let this metric be denoted by $J_{V}$.
\end{itemize}

\begin{figure*}[!t]
\centerline 
{
    \hspace{-1\baselineskip}
	\subfigure[$J_D$, 50 Nodes]
	{
		\includegraphics[width=\myvalc\columnwidth]{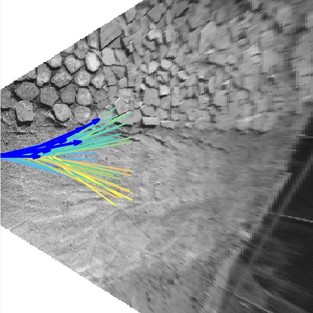}
		\label{fig:jd_50}
	}
	\hspace{-1\baselineskip}
	\subfigure[$J_D$, 100 Nodes]
	{
		\includegraphics[width=\myvalc\columnwidth]{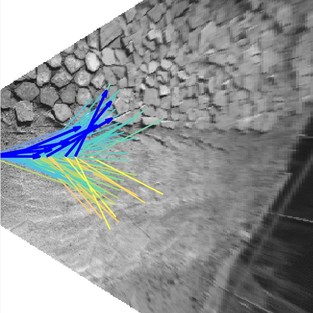}
		\label{fig:jd_100}
	}
	\hspace{-1\baselineskip}
	\subfigure[$J_D$, 250 Nodes]
	{
		\includegraphics[width=\myvalc\columnwidth]{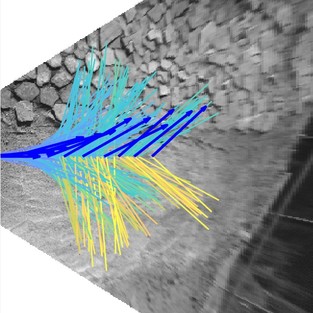}
		\label{fig:jd_250}
	}
	\hspace{-0.75\baselineskip}
	\subfigure[$J_D$, Optimal]
	{
		\includegraphics[height=\myvald]{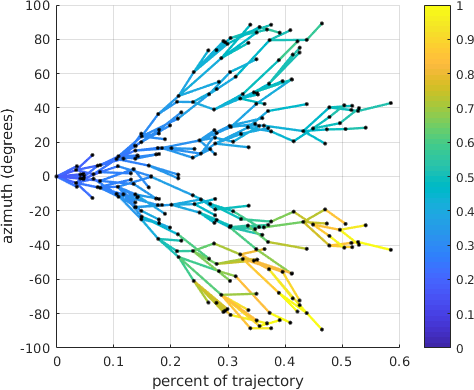}
		\label{fig:jd_tree}
	}
}
\centerline 
{
    \hspace{-1\baselineskip}
	\subfigure[$J_F$, 50 Nodes]
	{
		\includegraphics[width=\myvalc\columnwidth]{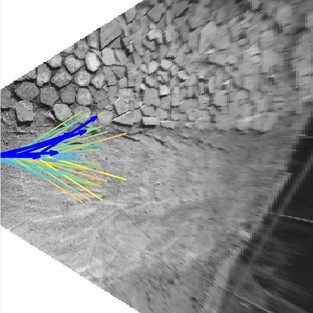}
		\label{fig:jf_50}
	}
	\hspace{-1\baselineskip}
	\subfigure[$J_F$, 100 Nodes]
	{
		\includegraphics[width=\myvalc\columnwidth]{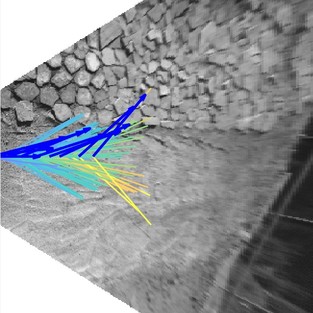}
		\label{fig:jf_100}
	}
	\hspace{-1\baselineskip}
	\subfigure[$J_F$, 250 Nodes]
	{
		\includegraphics[width=\myvalc\columnwidth]{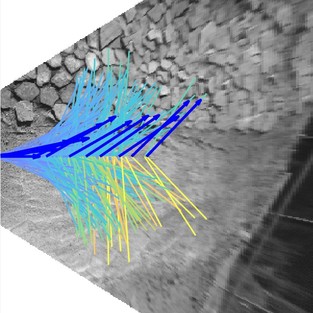}
		\label{fig:jf_250}
	}
	\hspace{-0.75\baselineskip}
	\subfigure[$J_F$, Optimal]
	{
		\includegraphics[height=\myvald]{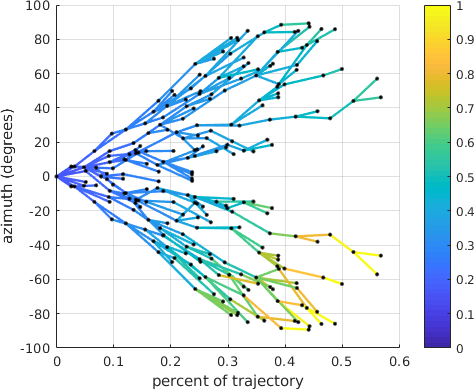}
		\label{fig:jf_tree}
	}
}
\centerline 
{
    \hspace{-1\baselineskip}
	\subfigure[$J_V$, 50 Nodes]
	{
		\includegraphics[width=\myvalc\columnwidth]{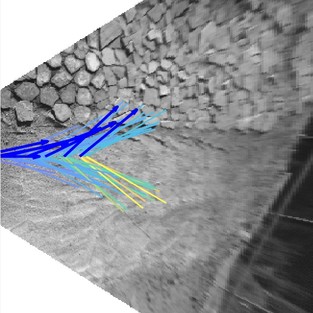}
		\label{fig:jv_50}
	}
	\hspace{-1\baselineskip}
	\subfigure[$J_V$, 100 Nodes]
	{
		\includegraphics[width=\myvalc\columnwidth]{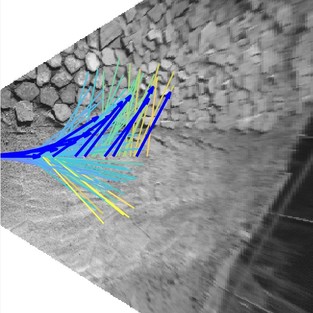}
		\label{fig:jv_100}
	}
	\hspace{-1\baselineskip}
	\subfigure[$J_V$, 250 Nodes]
	{
		\includegraphics[width=\myvalc\columnwidth]{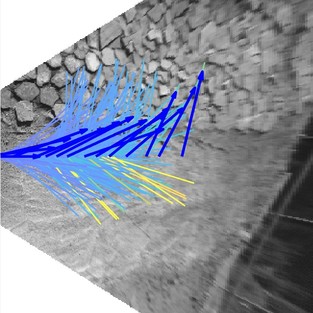}
		\label{fig:jv_250}
	}
	\hspace{-0.75\baselineskip}
	\subfigure[$J_V$, Optimal]
	{
		\includegraphics[height=\myvald]{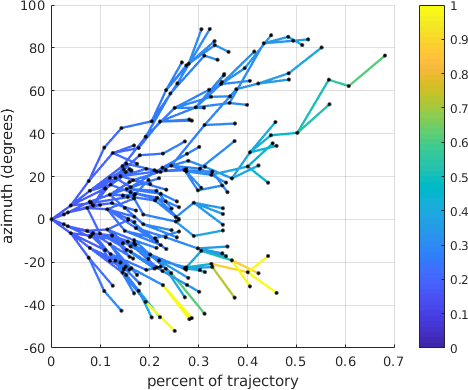}
		\label{fig:jv_tree}
	}
}
\caption{(Left) The optimal tree in the rover configuration space for 50 nodes, 100 nodes, and 250 nodes using VOSAP with each performance metric. \rev{The yellow edges indicate high cost and blue edges indicate low cost. The bold arrows represent the optimal mast trajectory. (Right) The optimal tree in the mast configuration space. The vertical axis is the pan angle in degrees (negative right, positive left), and the horizontal axis is the time represented as a percentage along the trajectory. Similarly, yellow edges indicate high cost and blue edges indicate low cost.}}
\label{fig:computation_reduction_tree}
\end{figure*}

The runtime for each performance indicator is reported in \autoref{tab:computation_reduction_runtime} for different tree sizes using VOSAP to compute the optimal trajectory. The steps of the algorithm are illustrated in \autoref{fig:computation_reduction_tree}. Notice the similarity between the optimal trajectory for each metric. The alternative performance metrics provide a similar trajectory for a significantly lower runtime; however, notice $J_{V} < J_{F} < J_{D}$ for samples pointing towards texture-poor terrain. This indicates that $J_{D}$ is a stronger indicator of performance, and in some cases, the alternative metrics may not converge to the same solution. This is expected since $J_{D}$ directly minimizes the localization drift; however, $J_{F}$ and $J_{V}$ provide an efficient alternative for quantifying the texture information at a significantly lower computational cost.

\begin{table}[!t]
    \caption{Runtime for VOSAP using alternative performance indicators}
	\centering
	\begin{tabular}{p{3.8cm}p{3.6cm}p{3.6cm}p{3.6cm}p{3.6cm}}
		\toprule
		& 50 Nodes & 100 Nodes & 250 Nodes \\
		\midrule
		~~~$J_{D}$ & 2.378\,s & 7.038\,s & 18.29\,s \\
		~~~$J_{F}$ & 0.148\,s & 0.340\,s & 0.978\,s \\
		~~~$J_{V}$ & 0.018\,s & 0.043\,s & 0.120\,s \\
		\bottomrule
	\end{tabular}
	\label{tab:computation_reduction_runtime}
\end{table}

\section{Conclusions} \label{sec:conclusion}

In this work, \rev{we employ a method} for predicting the performance of the perception system and incorporating \rev{the predicted performance} in the planning algorithm. 
The general problem of active perception is cast as a problem of planning under uncertainty. 
\rev{In our previous work \cite{Otsu18_wheretolook}, we proposed a method} for a VO-aware sampling-based planner (VOSAP) that actively seeks trajectories that improve the performance of perception leading to higher localization accuracy. 
To quantify the perception performance, VOSAP generates the most-likely future image measurements for candidate trajectories and evaluates the contribution of the perception system on localization performance. This paper focuses on the application of these methods to planetary surface exploration. \rev{The performance of the algorithm was demonstrated through a series of experiments in the Mars Yard at \ac{JPL} during different times of the day under various illumination conditions.} \rev{Additionally, alternative performance metrics were applied for improving the runtime of the overall algorithm.} 
In all experiments, the proposed method actively steers the perception system and outperformed the traditional fixed-mast setup, and in some cases, the proposed method avoided total failure in contrast to the passive configuration. 

\rev{Nevertheless, the proposed method is not without limitations, and in this regard, we discuss potential improvements and future work.
First, the performance gain of VOSAP over the benchmark is highly dependent on the texture and configuration of landscape. 
Second, the proposed method is valid only for systems that are capable of steering the visual sensors. 
This includes systems that are capable of actuating the visual sensors independent of the body, but the algorithm can easily be adapted to include systems with static sensors that are capable of following a path while varying attitude (e.g., holonomic robots and most \acp{MAV}). 
Third, the planning algorithm does not consider the coverage of the map in regards to images captured for \ac{VO}. As a result, additional images must be captured if the map become degenerate during runtime. 
The major drawback of this is that the degeneracy is detected only if the planning algorithm fails to plan a trajectory. 
This is not efficient since planning is performed unnecessarily in degenerate cases. 
In future work, either 1) map coverage should be incorporated in the planning algorithm to avoid degeneracy or 2) degeneracy should be detected prior to planning to avoid running the planner with a degenerate map.}
Finally, extending perception-aware methods to Mars helicopter-rover-orbiter coordination and navigation is an important future direction of our research \cite{ebadi2018aerial,nilsson2018guided,pflueger2019roverirl}.


\subsubsection*{Acknowledgments}

This research was supported in part by NASA Project WV-80NSSC17M0053. This research was carried out at the Jet Propulsion Laboratory, California Institute of Technology, under a contract with the National Aeronautics and Space Administration. Government sponsorship acknowledged.

\bibliographystyle{apalike}
\bibliography{references}

\end{document}